\documentclass[lettersize,journal]{IEEEtran}
\usepackage{amsmath,amsfonts,amssymb}
\usepackage{array}
\usepackage[caption=false,font=normalsize,labelfont=sf,textfont=sf]{subfig}
\usepackage{textcomp}
\usepackage{stfloats}
\usepackage{url}
\usepackage{verbatim}
\usepackage{cite}

\usepackage{hyperref}
\usepackage{tikz}
\usepackage{algorithm}
\usepackage{algpseudocode}
\usepackage{multirow}
\usepackage{graphicx}
\usepackage{booktabs}
\usepackage{placeins}
\usepackage{dsfont}
\usepackage[dvipsnames]{xcolor}
\usepackage{rotating}
\usepackage{wrapfig}
\usetikzlibrary{positioning}
\usetikzlibrary{arrows.meta}

\newtheorem{theorem}{Theorem}[section]

\newtheorem{proposition}[theorem]{Proposition}

\usepackage[table]{xcolor} 
\usepackage{pgf} 

\definecolor{myteal}{RGB}{94, 84, 65}

\usepackage{titlesec}

\let\oldparagraph\paragraph

\RenewDocumentCommand{\paragraph}{s m}{%
  \IfBooleanTF{#1}
    {\oldparagraph*{#2.}}%
    {\oldparagraph{#2.}}%
}

\titleformat{\paragraph}[runin]
  {\normalfont\normalsize\bfseries}{}{0pt}{}
\titlespacing*{\paragraph}
    {0pt}{1.5ex plus 0.5ex minus 0.2ex}{1em}

\newcommand{\GCCell}[2]{%
    \ifdim #1 pt > 0pt 
        \cellcolor{myteal!#1} #2 
    \else 
        #2 
    \fi 
}

\begin{document}

\title{Geometric Probing for Algorithm Selection in Continuous Black-Box Optimization}


\author{
Jiabao Brad Wang,
Xiang Shi,
Yiliang Yuan,
and Mustafa M{\i}s{\i}r%
\thanks{Jiabao Brad Wang, Xiang Shi, and Mustafa M{\i}s{\i}r are with Duke Kunshan University, Kunshan, Jiangsu, China 
(e-mail: jb.wang@dukekunshan.edu.cn; xiang.shi@dukekunshan.edu.cn; mustafa.misir@dukekunshan.edu.cn).}%
\thanks{Yiliang Yuan is with Mohamed bin Zayed University of Artificial Intelligence, Abu Dhabi, United Arab Emirates 
(e-mail: yiliang.yuan@mbzuai.ac.ae).}%
\thanks{This work has been submitted to the IEEE for possible publication. Copyright may be transferred without notice, after which this version may no longer be accessible.}
}



\maketitle

\begin{abstract}
Automated algorithm selection for continuous black-box optimization depends on what information is acquired from a problem under a limited probing budget and how that information is represented. We introduce a geometric probing framework that samples multi-scale two-dimensional restrictions across location, orientation, and scale, and encodes their normalized objective-value maps with validity-aware convolutional processing and permutation-invariant aggregation. We compare our method with classical ELA and Deep-ELA under matched budgets, within-problem and problem-level transfer, fusion, representation, and budget analyses. We further disentangle probe acquisition from probe processing by controlled ablation. The results show that the proposed visual representation exposes solver-performance information complementary to ELA-family features and retains a relative advantage for relative expected runtime under problem-level transfer, while greater coverage, feature richness, or predictive accessibility alone does not guarantee better selection.
\end{abstract}


\section*{Code availability}
The implementation is available in an archival repository at \url{https://doi.org/10.5281/zenodo.22822605}.

\section{Introduction}
\label{sec:introduction}

Continuous black-box optimization (BBO) concerns problems accessible only through objective evaluations, without analytic expressions or gradients and sometimes under expensive or noisy evaluations. 
Since gradients are unavailable, the standard approach is to use derivative-free and stochastic optimizers, whose relative performance varies substantially with properties including conditioning, modality, separability, and local geometry~\cite{kerschke2019automated}.

Motivated by this variability, automated algorithm selection (AS) \cite{kerschke2019automated} chooses a solver which is expected to perform the best for each problem instance.
Static AS performs selection once prior to the optimization on the target instance begins.
Since the objective is unknown and only a limited probing budget is available, static AS first characterizes the instance following a representation scheme.
This entails two linked choices: a probing operator determines which evaluations and relations are observed, and a processor determines what information is retained in the representation. Most established pipelines combine space-filling point designs with handcrafted or learned landscape descriptors~\cite{kerschke2019automated,cenikj2026survey,seiler2025deep}. They perform effectively when training and test cases share problem structure, yet often deteriorate under problem-level or cross-benchmark transfer~\cite{dietrich2024impact,petelin2025pitfalls,cenikj2025landscape}. This raises a question preceding model capacity: which probe--processor biases expose solver-relevant structure that remains useful when the problem family changes?

We study this question through \emph{geometric probing}. We propose a probe-processor pipeline which samples coarse two-dimensional restrictions across location, orientation, and logarithmic scale in the normalized search domain. A shared convolutional processor encodes each normalized value map with its validity mask, conditions the encoding on scale and amplitude statistics, and averages the resulting slice representations. 

Our analysis treats acquisition and processing as separable, interacting design choices. We compare the canonical slice/visual and Sobol/ELA-family pipelines, apply several processors to shared probes, examine fusion and representation complementarity, and vary slice geometry and probe budget. BBOB experiments show that our method supplies complementary information and becomes relatively stronger under problem-level shift, especially for typical relERT outcomes. They also delimit this advantage that additional observations and more accessible absolute-performance information do not necessarily yield safer solver choices.

The contributions are:
\begin{enumerate}
    \item a structured multi-scale 2-d restriction probe and validity-aware visual processor for static AS in box-constrained BBO; 
    \item an analysis of slices' boundary validity and its dimensional scaling;
    \item a controlled empirical comparison with common ELA-family descriptors and decomposition of acquisition, processing, complementarity, predictive accessibility, and budget effects, evaluated under within-problem and problem-level generalization.
\end{enumerate}

The remainder of the paper is organized as follows. \S\ref{sec:background} reviews probe acquisition, probe processing, and generalization in static AS. \S\ref{sec:Methods} and \S\ref{sec:slice-analytics} introduce our methodology and analyze the geometry of slice validity. \S\ref{sec:experimental-design} specifies the benchmarks, targets, and experimental design; \S\ref{sec:results} reports the empirical analyses; and \S\ref{sec:discussion} discusses their implications and limitations.

\paragraph*{Extension over the conference version}
An earlier conference version introduced the slice representation and an initial 12-solver BBOB study with problem-level relERT labels, attention pooling, and a tail-aware composite selection score~\cite{wang2026geometric}. The present work changes \emph{both scope and evidence}. It uses the simpler mean-pooling-based encoder and a common MLP selection pipeline, so the comparison centers on representation. It moves to instance-level targets of an additional anytime measure AOCC, with 95 more instances per BBOB function; separates acquisition from processing; and adds matched-cost fusion, representation, and nested-budget analyses. The geometric treatment of slice validity is also new. The resulting study examines where geometric probing gains arise and where its representation--decision relation fails to generalize.

\section{Background}
\label{sec:background}

AS for BBO typically involves (i) extracting problem-instance information, (ii) collecting solver performance data, and (iii) training a predictive model to select a solver for an unseen problem~\cite{alissa2023automated}. 
In the broader MetaBBO view, AS is one component of automated algorithm design for black-box optimization, alongside algorithm configuration, solution manipulation, and algorithm generation~\cite{ma2025toward}. 
A widely used static pipeline is probe-then-predict: the optimization problem $f$ is queried through a probing operator $\mathcal P$ under a budget $B$, producing observations $\mathcal P(f)$, and a processing or representation map $\Phi$ transforms the probe into a representation $H=\Phi(\mathcal P(f))$ used by a downstream selector. 
Although $\mathcal P$ and $\Phi$ are conceptually distinct, they are usually instantiated and evaluated as a \emph{coupled} problem-representation pipeline~\cite{kerschke2019automated,cenikj2026survey}.

In continuous black-box optimization, probing has most commonly relied on generic space-filling point designs, including random, Latin-hypercube, and low-discrepancy samples such as Sobol sequences~\cite{owen1998scrambling,vskvorc2021effect,renau2020exploratory}, followed by a processor that summarizes the resulting evaluations. 
The canonical example is \textbf{Exploratory Landscape Analysis} (ELA), which estimates properties such as modality, curvature, separability, and conditioning from a limited set of sampled evaluations~\cite{kerschke2019automated,mersmann2011exploratory,prager2024pflacco}. Information-content landscape analysis provides an important related line, showing that compact descriptors extracted from sampled continuous landscapes can support problem characterization and downstream learning tasks~\cite{munoz2014exploratory}. These methods established that limited probes can contain useful solver-relevant information and remain the principal reference family for static AS~\cite{kerschke2019automated}. 
At the same time, the probe is not neutral to the representation: ELA features can change materially with the sampling strategy~\cite{renau2020exploratory}, and the sampling design can affect their invariance to function transformations~\cite{vskvorc2021effect}. 
Thus even for conventional point-set representations, what is sampled and how it is processed jointly determine the information exposed to the selector.

Subsequent work has diversified the processing side substantially. 
Topological landscape analysis summarizes sampled landscapes through layered persistence images~\cite{petelin2024tinytla}, while learned representations include VAE-based latent encodings in DoE2Vec~\cite{van2023doe2vec}, transformer-based selection in TransOptAS~\cite{cenikj2024transoptas}, and self-supervised transformer representations in Deep-ELA~\cite{seiler2025deep}. 
Controlled comparisons show that learned representations can be competitive with classical ELA under matched supervision and evaluation protocols~\cite{prager2022automated}, and combinations of learned and handcrafted descriptors can exploit complementary information~\cite{seiler2024learned,seiler2024synergies,gjorgjieva2026structural}. 
Most of this development changes how available evaluations are encoded while retaining a conventional point-sampling view of the landscape. 
A closely related departure is a visual-representation approach which evaluates random two-dimensional subspaces and renders the resulting observations as contour plots for solver-performance learning~\cite{yuan2026beyondarxiv}.
Here the acquisition and processor are jointly structured around spatial landscape information, providing evidence that preserving the arrangement of sampled values can be useful beyond handcrafted summary descriptors.

Generalization to unseen problems has been an explicit concern in AS for continuous BBO, motivating evaluation settings beyond instance-based or random within-suite splits~\cite{bischl2012algorithm,kerschke2019automated1}. Selectors based on ELA features or learned static representations are known to degrade under stronger distribution shifts, including problem-split and cross-benchmark evaluation, and can approach single-best-solver behavior~\cite{nikolikj2024generalization,petelin2024generalization,cenikj2026survey}. Consequently, gains under within-suite validation need not imply transfer to unseen problem types. Existing analyses further suggest that increasing model capacity or feature richness alone does not reliably remove this degradation~\cite{cenikj2025landscape,cenikj2025recent}. 

In this context, the present work explores the representation itself, including both the observations acquired and the invariances or relations retained by their processor. In particular, we couple a structured multi-scale slice probe with a visual processor specialized to the resulting local restrictions, and the experiments subsequently vary acquisition and processing separately as well as jointly.
This provides a controlled setting for disentangling how much solver-relevant information arises from where the probing budget is spent, from how those observations are represented, or from the pairing between the two.

\section{Methodology}
\label{sec:Methods}

\subsection{Geometric Probing}
\label{subsec:geometric-probe}
\begin{wrapfigure}{r}{0.41\linewidth}
    \includegraphics[width=\linewidth]{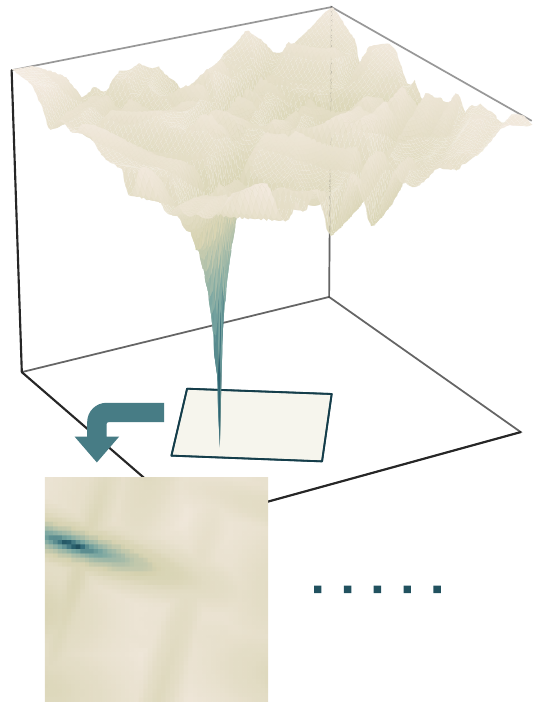}
\end{wrapfigure}
Given a box-constrained optimization problem instance $f$, we seek a probe operator $\mathcal P$ of its local landscape geometry that can reveal solver-relevant patterns such as anisotropy, basin shape, oscillation, and boundary interaction without committing to a fixed global coordinate system. 
To this end, we represent $f$ by a finite set of low-resolution two-dimensional \emph{restrictions} sampled across location, orientation, and scale, under the hypothesis that this collection preserves useful local geometric information. 

\begin{wrapfigure}{l}{0.4\linewidth}
    \includegraphics[width=\linewidth]{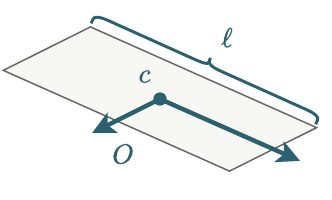}
\end{wrapfigure}
Each restriction is an \emph{oriented square slice} of the normalized search domain $[0,1]^d$, parameterized by a triple $(c, O, \ell)$, where $c \in [0,1]^d$ is the slice center, $O \in \mathbb{R}^{d \times 2}$ is the orientation satisfying $O^\top O = I_2$, and $\ell > 0$ is the side length, i.e., the slice scale.

\paragraph*{Sampling variables}
We want the sampled variables $(c, O, \ell)$ to satisfy three principles under a finite probing budget: reasonably even coverage of the domain, no a priori preference for particular directions, and no a priori preference for particular absolute scales.
Centers $c$ are sampled using a scrambled Sobol sequence~\cite{owen1998scrambling} in $[0,1]^d$ to obtain more even coverage than i.i.d. uniform sampling under limited budget~\cite{vskvorc2021effect}.
The orientation sampling is chosen to be orthogonally invariant to avoid preferring particular two-dimensional subspaces of $\mathbb R^d$, nor particular orthonormal coordinate systems within a sampled subspace. 
In practice, this is achieved by drawing a Gaussian matrix $G \in \mathbb{R}^{d \times 2}$ with i.i.d. standard normal entries and orthonormalizing its columns.
The resulting frame is uniformly distributed on the Stiefel manifold
$$O\sim\mathcal U(\operatorname{St}(d, 2)), \qquad\operatorname{St}(d, 2) = \left\{O\in\mathbb R^{d\times2}: O^TO=I_2\right\}$$
and the induced distribution on $\mathrm{span}(O)$ is the uniform invariant distribution over two-dimensional subspaces~\cite{mezzadri2006generate}.
Scales $\ell$ are sampled independently from a log-uniform distribution,
\[
\log \ell \sim \mathcal{U}(\log \ell_{\min}, \log \ell_{\max}),
\]
which reflects multiplicative scale neutrality that equal relative changes in slice size should be treated comparably, so equal probability mass is assigned to equal intervals in $\log \ell$.
Throughout, $\ell$ is defined relative to $[0,1]^d$ and is the Euclidean side length of the slice.
We use $\ell_{\min}=0.02$ and $\ell_{\max}=1/\sqrt{2}$. The lower bound avoids numerically degenerate slices at fixed raster resolution, while the upper bound is an intuitive fixed-value baseline which is the maximum value that guarantees no boundary truncation when $C$ is at the center of $[0, 1]^d$.
In practice, $c$ and an auxiliary scale coordinate are sampled jointly using a scrambled Sobol sequence in $[0, 1]^{d+1}$, and the auxiliary coordinate is mapped through the inverse log-uniform transform, yielding the marginal design. 

\begin{figure}[t]
    \centering
    \includegraphics[width=\linewidth]{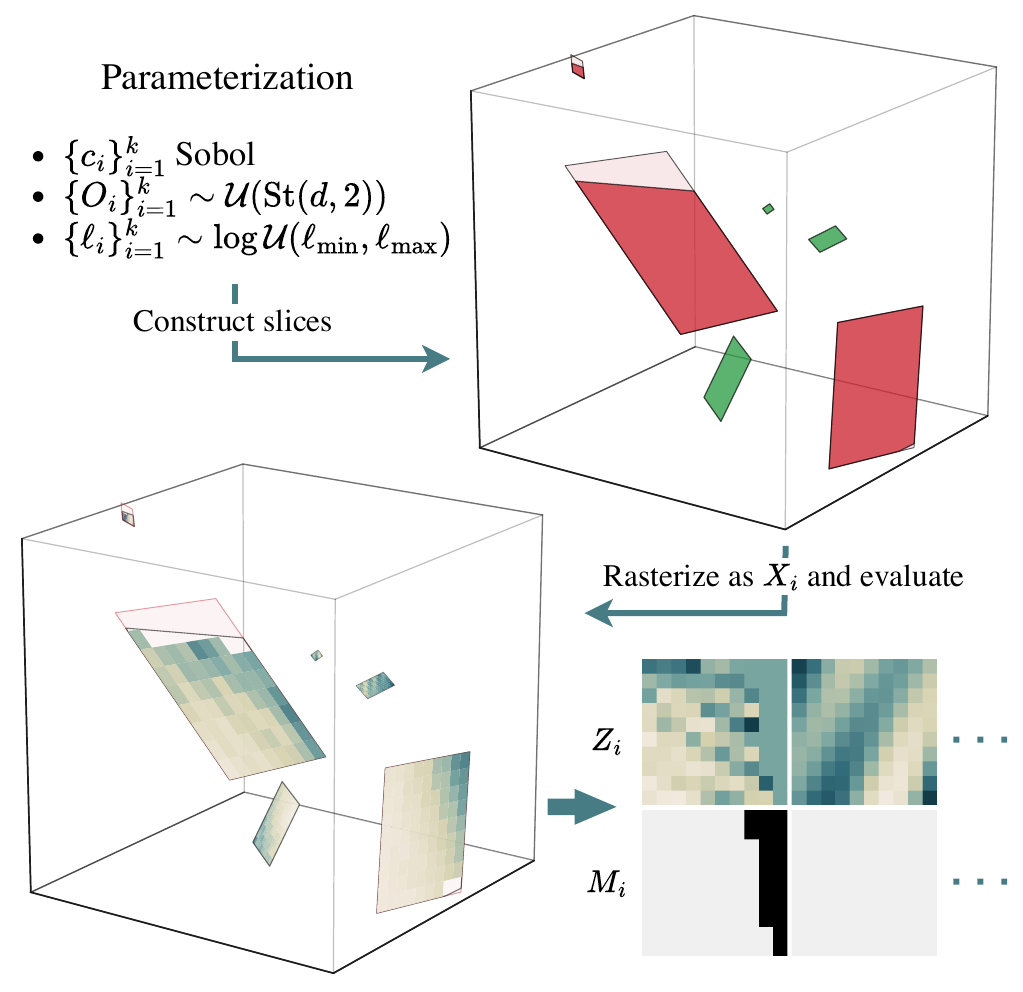}
    \caption{Illustration of the proposed slice-based probing strategy $\mathcal P_{\mathrm{slice}}$ in the normalized domain $[0,1]^d$. The cubes demonstrate an example of $d = 3$. Green slices are entirely within the domain and red ones are partially outside. 
}
    \label{fig:plot_mask_demo}
\end{figure}

\paragraph*{Slice construction}
Given $(c, O, \ell)$, we define the local continuous slice map
\begin{equation}\label{eq:local-slice-map}
    x(u) = c + \ell O u, \qquad u \in \left[-\tfrac{1}{2}, \tfrac{1}{2}\right]^2.
\end{equation}
We discretize the local coordinates $u$ on a regular $r \times r$ grid $\left\{u^{(a,b)}\right\}_{a,b=1}^r$, map the resulting points back to the original problem domain, e.g. $[-5,5]^d$ for BBOB, and evaluate $f$ there. 
Grid points that fall outside the problem domain are marked invalid and are not evaluated. Their coordinates are clipped to the nearest feasible point only to preserve a dense $r\times r$ tensor representation.
These entries are explicitly marked invalid and are excluded from all subsequent summary statistics.
This results in a map $Z\in\mathbb R^{r\times r}$ of function values. 
To preserve the semantics of hard domain constraints, we also store a binary validity mask \(M \in \{0,1\}^{r \times r}\) with
\[
    M^{(a,b)} = \mathds{1}\left[x\left(u^{(a,b)}\right) \in [0,1]^d\right],
\]
which is intended to make boundary truncation explicit instead of treating it as ordinary landscape structure by excluding invalid regions later during feature extraction. 
For a set of $k$ slices (regarding $k$ values, see \S\ref{subsec:representation-panel}), the full probe set is thus 
\begin{equation}\label{eq:full-probe}
    \mathcal P _{\mathrm{slice}}(f) = \left\{(c_i, O_i, \ell_i, X_i, Z_i, M_i)\right\}_{i=1}^k, 
\end{equation}
an illustration of which is given in Figure~\ref{fig:plot_mask_demo}. 
The suite of coordinate and evaluation maps $(X_i, Z_i)$ can be directly used for objective-only landscape analysis like ELA via pulling only the valid points, whereas next we provide a pipeline specialized for processing the proposed slice-structured probes. 

\begin{figure*}[t]
    \centering
    \includegraphics[width=\linewidth]{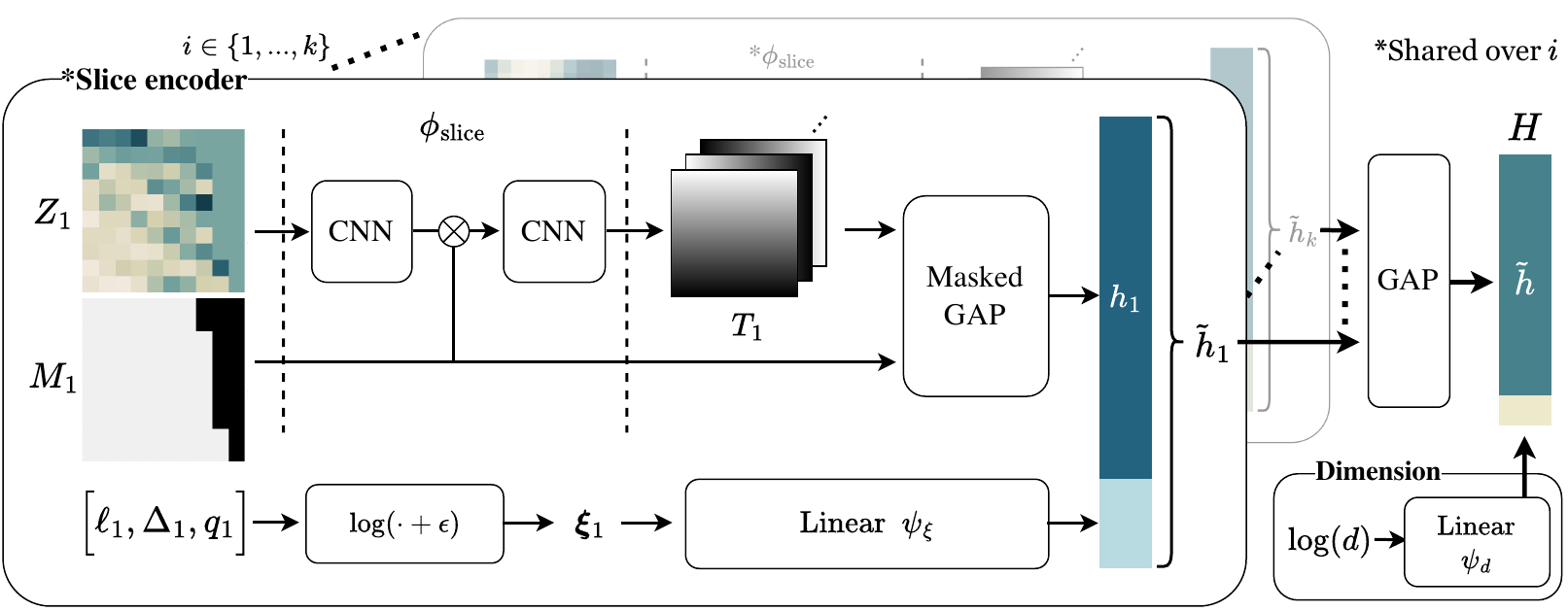}
    \caption{Overview of the visual encoder $\Phi_{\mathrm{visual}}$.
Each sampled slice contributes a value map $Z_i$, validity mask $M_i$, and side statistics $(\ell_i,\Delta_i,q_i)$.
A shared visual encoder $\phi_{\mathrm{slice}}$ processes $Z_i$ with a convolutional network, with the mask multiplied between convolutions to zero invalid locations and then used in masked GAP to produce a slice embedding $h_i$.
After concatenation with an embedding of the log-transformed side statistics, the conditioned slice features $\tilde h_i$ are aggregated by average pooling, and a low-capacity embedding of $\log d$ is appended to form the final instance representation $H$.}
    \label{fig:encoder_demo}
\end{figure*}

\subsection{Visual Encoder}
\label{subsec:encoder}

Our probe-processing map $\Phi_{\mathrm{visual}}:\mathcal P(f)\mapsto H$ is designed to 1) learn per-slice geometry, 2) exclude invalid function values while retaining boundary-validity information, 3) reduce sensitivity to arbitrary function-value scale, and 4) avoid dependence on the absolute position and orientation of slices in the search space. It should be noted that these are not canonical preferences for AS, but design choices motivated respectively by our representation hypothesis, the semantics of box-constrained optimization, empirical evidence that removing nuisance transformations can improve invariance and generalization~\cite{prager2023nullifying}, and the use of translations and rotations in generating BBOB instances. The first three are reflected in the preprocessing and masked encoding pipeline, whereas the last is encouraged by excluding the probed spatial locations $X_i$ and their positional parameterizations $(c_i, O_i)$ from the encoder input. An overview of $\Phi_{\mathrm{visual}}$ is given in Figure~\ref{fig:encoder_demo}.

\paragraph*{Preprocessing} Given a probe set, we retain only $\left\{(Z_i, M_i, \ell_i)\right\}_{i=1}^k$ and disregard the probed spatial locations as well as their positional parameterization, $\left\{(X_i, c_i, O_i)\right\}_{i=1}^k$. The valid function values are first min-max normalized slice-jointly over the probe to remove arbitrary probe-level offset and amplitude. For shared-CNN processing, the evaluation map $Z_i$ of each slice is further individually min-max normalized to $[-1/2,1/2]$, where invalid entries are assigned zero to stay neutral as well as to reduce their effects in the first downstream convolution, resulting in a normalized map $\hat Z_i\in [-1/2, 1/2]^{r\times r}$. 
This prevents boundary truncation from being confounded with flat or low-variance landscapes and removes arbitrary per-slice offsets and amplitudes, so that the image encoder primarily models within-slice geometry. 
To avoid discarding inter-slice amplitude discrimination \textit{within} an instance altogether, we record, from the globally normalized values but before per-slice normalization, the range $\Delta_i$ and the interquartile range (IQR) $q_i$ of each slice for later conditioning. 


\paragraph*{Per-slice encoding and masked spatial pooling}
Each slice is processed independently by a shared encoder
\begin{equation}
    \phi_{\mathrm{slice}}:\left(\hat Z_i,M_i\right)
\mapsto
T_i
\in\mathbb{R}^{C_Z\times r\times r},
\end{equation}
where $\phi_{\text{slice}}$ comprises two lightweight convolutional blocks without spatial downsampling, and the validity mask is reapplied to every feature channel to zeroize the invalid positions between the two blocks. 
The final feature tensor is reduced by masked global average pooling (GAP), 
\begin{equation}
    h_i
=
\frac{1}{\sum_{a,b=1}^rM_i^{(a,b)}}
\sum_{a,b=1}^r M_i^{(a,b)} T_i^{(:,a,b)}
\in\mathbb{R}^{C_Z}.
\end{equation}
Thus invalid locations carry no objective-value activation into the convolutions nor contribute directly to the masked GAP.

\noindent
\paragraph*{Scale and amplitude conditioning}
To disambiguate per-slice geometry across slice scales and value magnitudes, we reintroduce a small side channel
\[
\boldsymbol{\xi}_i
=
\bigl(
\log \ell_i,\,
\log(\Delta_i+\varepsilon_\xi),\,
\log(q_i+\varepsilon_\xi)
\bigr)\in\mathbb{R}^3,
\]
where \(\varepsilon_\xi=10^{-6}\).
The logarithm places these quantities on a more comparable numerical scale across orders of magnitude.
We embed \(\boldsymbol{\xi}_i\) by a one-layer map \(\psi_\xi:\mathbb{R}^3\to\mathbb{R}^{C_\xi}\)
for $C_\xi\in\mathbb Z^+$ and concatenate it with the visual embedding:
\[
\tilde h_i
=
\bigl[h_i \mid \psi_\xi(\boldsymbol{\xi}_i)\bigr]
\in\mathbb{R}^{C_Z+C_\xi}.
\]

\paragraph*{Permutation-invariant slice aggregation}
The set of conditioned slice-level embeddings \(\{\tilde h_i \}_{i=1}^k\) is aggregated permutation-invariantly into an instance-level representation by average pooling
$
\bar h
=
\frac1k\sum_{i=1}^k\tilde h_i \in\mathbb{R}^{C_Z+C_\xi}
$.


\paragraph*{Ambient-dimensionality conditioning}
Finally, we append an embedding of the ambient dimension, also being instance-level, learned by a linear layer \(\psi_d:\mathbb{R}\to\mathbb{R}^{C_d}\) for $C_d\in\mathbb Z^+$ and form the final instance representation
\[
H = \bigl[\bar h \mid \psi_d(\log d)\bigr]\in\mathbb{R}^{C_Z+C_\xi+C_d}.
\]
This provides a low-capacity amount of global context, since similar local slice geometry need not imply the same algorithmic behavior across different ambient dimensions. 
The feature widths are empirically configured to be $C_Z=128, C_\xi = 16, C_d = 4$.

\begin{algorithm}[t]
\caption{GeoPAS}
\label{alg:geopas}
\begin{algorithmic}[1]
\Require Problem instance $f$, dimension $d$, number of slices $k$, resolution $r$, scale range $[\ell_{\min},\ell_{\max}]$
\Ensure Selected algorithm $\hat a$

\State $\operatorname{GeoPAS} = \operatorname{Select} \circ \ \Phi_{\mathrm{visual}} \circ \mathcal P_{\mathrm{slice}} $

\Statex

\Function{$\mathcal P_{\mathrm{slice}}$}{$f$}
    \State Sample centers $\{c_i\}_{i=1}^k$ by scrambled Sobol design
    \For{$i=1,\ldots,k$}
        \State Sample
        $O_i\sim\mathcal U(\operatorname{St}(d,2))$ and
        $\log\ell_i\sim
        \mathcal U(\log\ell_{\min},\log\ell_{\max})$
        \State Discretize
        $x_i(u)=c_i+\ell_iO_i u$
        and evaluate $f$
        \State Record $(X_i,Z_i,M_i)$
    \EndFor
    \State \Return
    $\{(c_i,O_i,\ell_i,X_i,Z_i,M_i)\}_{i=1}^k$
\EndFunction

\Statex

\Function{$\Phi_{\mathrm{visual}} $}{$\mathcal P$}
    \State Jointly normalize valid probe values
    \For{$i=1,\ldots,k$}
        \State Compute $(\Delta_i,q_i)$ and obtain $\hat Z_i$
        \State
        $h_i\gets
        \operatorname{MaskedGAP}
        \!\left(\phi_{\mathrm{slice}}(\hat Z_i,M_i),M_i\right)$
        \State
        $\tilde h_i\gets
        [\,h_i\mid\psi_\xi(\log\ell_i,
        \log(\Delta_i+\varepsilon_\xi),
        \log(q_i+\varepsilon_\xi))\,]$
    \EndFor
    \State
    $\bar h\gets k^{-1}\sum_{i=1}^k\tilde h_i$
    \State
    $H\gets[\bar h\mid\psi_d(\log d)]$
    \State \Return $H$
\EndFunction

\Statex

\Function{$\operatorname{Sel}$}{$H$}
    \State $\hat{\boldsymbol y}\gets h_\theta(H)$
    \State $\hat a\gets
    \operatorname*{arg\,opt}_{a\in\mathcal A}\hat y_a$
    \State \Return $\hat a$
\EndFunction

\end{algorithmic}
\end{algorithm}

We combine the geometric probing operator $\mathcal P_{\mathrm{slice}}$, the visual processing map $\Phi_{\mathrm{visual}}$, and the downstream MLP-based selector $\operatorname{Sel}: H\mapsto \hat a\in\mathcal A$ (see \S\ref{subsec:evaluation-pipeline}) into an end-to-end AS pipeline. For brevity, we refer to this composition as \textbf{geometric probing-based algorithm selection (GeoPAS)}:
\begin{equation}
    \operatorname{GeoPAS} = \operatorname{Sel} \circ\ \Phi_{\mathrm{visual}} \circ \mathcal P_{\mathrm{slice}},
\end{equation}
a schematic of which is given in Algorithm~\ref{alg:geopas}. 

\section{Slice analytics}
\label{sec:slice-analytics}

The proposed slice probing excludes points outside the box search domain. This truncation may scale with dimensionality, as uniformly distributed centers tend to be close to the box boundary in high-dimensional space. Indeed, let $d\geq2$, $Q_d = [0,1]^d$ denote the unit hypercube and $C\sim\mathcal U(Q_d)$, then for any fixed $\epsilon\in(0, 1/2)$,
$$
\mathbb P\left(\operatorname{dist}(C, \partial Q_d) > \epsilon\right) = (1 - 2\epsilon)^d \longrightarrow 0
$$
as $d$ increases. Next we analyze how the whole slices do.

\noindent
\textit{Expected valid fraction.}
Let $S = \left[-\frac{1}{2}, \frac{1}{2}\right]^2$, and we consider the randomized form of the continuous slice map~\ref{eq:local-slice-map}, 
\begin{equation}\label{generalised-slice-map}
    X(u) = C + L \mathcal O u,
\qquad u\in S,
\end{equation}
where $C\sim\mathcal U(Q_d)$ describes the uniform-measure average that the Sobol center design is intended to approximate, $\mathcal O\sim \mathcal U(\operatorname{St}(d, 2))$, and $\log L\sim\mathcal{U}(\log \ell_{\min}, \log \ell_{\max})$ with $0<\ell_{\min}<\ell_{\max}$. We further let $U\sim\mathcal U(S)$ represent a uniformly sampled point within the slice.

Since $S$ has unit area, averaging the validity of a uniformly sampled $U$ gives the valid-area fraction of a realized slice. Consequently, its expectation over centers, orientations and scales is
\begin{equation}
    \bar\eta_d(\ell_{\min}, \ell_{\max}) := 
    \mathbb P(C + L \mathcal O U\in Q_d).
\end{equation}
We first observe that, for any fixed $\ell$ and $u\neq0$, the normalized direction of the displacement is uniformly distributed on the unit sphere $\mathbb S^{d-1}$
$$
\frac{l\mathcal O u}{\|l\mathcal O u\|_2} = \frac{\mathcal O u}{\|u\|_2} \sim \mathcal U\left(\mathbb S^{d-1}\right),
$$
as $\mathcal O$ is rotationally invariant. We thereby fix values $L = \ell$ and $U = u \neq0$, let $\sigma = \|\ell Ou\|_2 = \ell\|u\|_2$ be the displacement's length and $V = \mathcal Ou/\|u\|_2 \sim \mathcal U\left(\mathbb S^{d-1}\right)$ be its normalized direction, and define
\begin{equation}
    p_d(\sigma) = \mathbb P(C + \sigma V\in Q_d) \equiv \mathbb P(C + l \mathcal O u\in Q_d\mid \ell, u)
\end{equation}
to be the conditional probability, where $C$ and $V$ are independent.  This also gives the relation
\begin{equation}\label{eq:sa-radial-average}
    \bar\eta_d(\ell_{\min},\ell_{\max})
    =\mathbb E_{L,U}\bigl[p_d(L\|U\|_2)\bigr].
\end{equation}

For a fixed displacement $\sigma V$, coordinate $i$ remains in $[0,1]$ with probability $\max\left(\left (1 - \sigma|V_i|\right), 0\right)$. Independence of the coordinates of $C$ therefore gives
\begin{equation}
    p_d(\sigma) = \mathbb E_V\left[\prod_{i=1}^d\max\left(\left (1 - \sigma|V_i|\right), 0\right)\right].
\end{equation}
Since $\|U\|_2\leq1/\sqrt{2}$ and $|V_i|\leq 1$, we have $1 - \sigma|V_i|\geq 1-1/\sqrt{2}L > 0$ if and only if $L<\sqrt{2}$. We assume so as this is already higher than the axis length $1$, and therefore
\begin{equation}\label{eq:sa-product}
    p_d(\sigma) = \mathbb E_V\left[\prod_{i=1}^d\left (1 - \sigma|V_i|\right)\right].
\end{equation}
Expanding the product and using exchangeability of the sphere coordinates gives the following exact expectation.
\begin{proposition}[Expected continuous valid fraction]
\label{prop:sa-exact}
Under the preceding assumptions,
\begin{equation}\label{eq:sa-exact-fraction}
    \bar\eta_d
    =\sum_{m=0}^d(-1)^m\binom dm
    \frac{\Gamma(d/2)}
    {\pi^{m/2}\Gamma((d+m)/2)}\cdot\mathbb E\left[L^m\right]\cdot \mathbb E\left[\|U\|_2^m\right],
\end{equation}
where $\mathbb E\left[L^0\right] = \mathbb E\left[\|U\|_2^0\right]=1$, and, for $m\ge1$,
\begin{align}
    \mathbb E\left[L^m\right]&=\frac{\ell_{\max}^m-\ell_{\min}^m}
    {m\ln(\ell_{\max}/\ell_{\min})},\label{eq:sa-scale-moment}\\
    \mathbb E\left[\|U\|_2^m\right]&=\frac{2^{1-m}}{m+2}
    \int_0^{\frac\pi4}\sec^{m+2}\theta\,d\theta.
    \label{eq:sa-square-moment}
\end{align}
\end{proposition}
\begin{IEEEproof}
See Appendix~\ref{app:sa-exact}.
\end{IEEEproof}


\paragraph*{Limiting dimension scaling}
The exact expectation permits finite-dimensional evaluation, and its dependence on $d$ is clearer from the product representation above. 
Let
\[
q_d(V;\sigma)
:=
\prod_{i=1}^d\max\{1-\sigma|V_i|,0\},
\qquad
p_d(\sigma)=\mathbb E_V[q_d(V;\sigma)].
\]
For $\sigma_d=t/\sqrt d$, Appendix~A2 shows that
\[
q_d(V;\sigma_d)
\xrightarrow{\mathbb P}
\exp\!\left(-t\sqrt{\frac{2}{\pi}}\right).
\]
Since $0\leq q_d\leq1$, its expectation converges to the same limit. Moreover,
\[
\mathbb E\lVert V\rVert_1
=
\frac{d\,\Gamma(d/2)}{\sqrt\pi\,\Gamma((d+1)/2)}
\sim
\sqrt{\frac{2d}{\pi}}.
\]
The accumulated coordinate displacement is therefore governed by $\sigma\sqrt d$, motivating a scale of order $1/\sqrt d$. More precisely, we obtain the following limit.

\begin{proposition}[Limiting dimension scaling]
\label{prop:sa-exact}
For any constant $t>0$,
\begin{equation}\label{eq:sa-critical-limit}
    \lim_{d\rightarrow\infty}p_d\left(\frac{t}{\sqrt d}\right)
    \longrightarrow \exp\left(-t\sqrt{\frac2\pi}\right). 
\end{equation}
In contrast, for fixed positive scale endpoints, $\bar\eta_d(\ell_{\min},\ell_{\max})\to0$.
\end{proposition}
\begin{IEEEproof}
See Appendix~\ref{app:sa-scaling}.
\end{IEEEproof}
These results motivate comparing the original $\ell_{\max}=1/\sqrt2$ with a dimension-adaptive maximum $\ell_{\max}=1/\sqrt d$. The latter matches the original maximum at $d=2$.

\begin{figure}[t]
    \centering
    \includegraphics[width=0.75\linewidth]
    {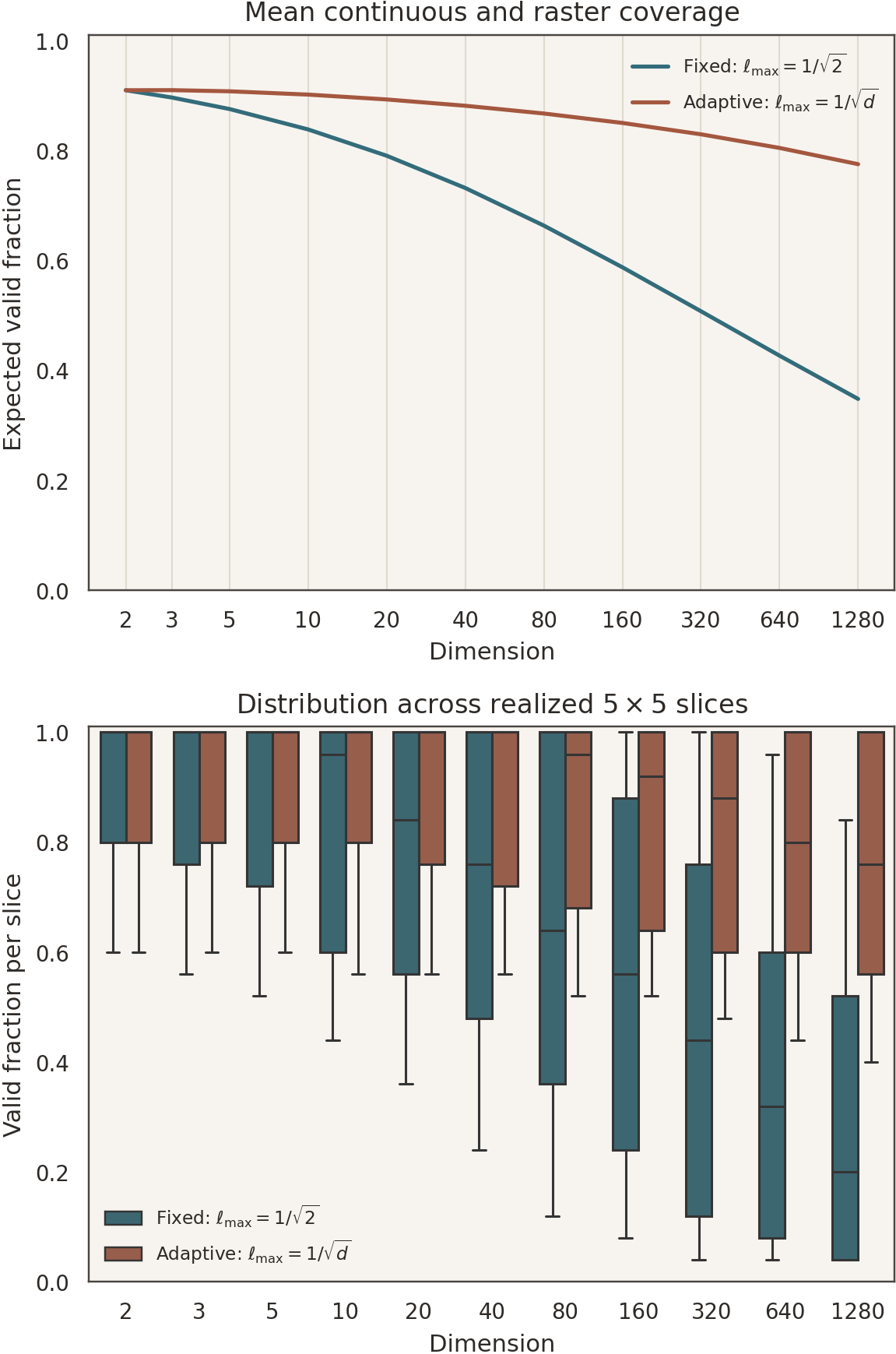}
    \caption{Boundary validity, under the two scale designs, with $\ell_{\min}=0.02$ and either $\ell_{\max}=1/\sqrt2$ or $1/\sqrt d$. (Top) the expected continuous valid fraction and (bottom) the distribution of realized $5\times5$ raster fractions.}    
    \label{fig:sa-fixed-min}
\end{figure}

\paragraph*{Numerical illustration}
We compare the numerically computed expected continuous valid fractions \ref{eq:sa-exact-fraction} and realized raster valid fractions with $8{,}192$ slices sampled, both over $d\in[2, 1280]$. Figure~\ref{fig:sa-fixed-min} compares the two maximum-scale designs with $\ell_{\min}=0.02$. The expected continuous valid fraction decreases with $d$ for both designs over the displayed dimensions, with a substantially smaller decrease under the adaptive maximum. The realized raster fractions likewise retain more valid points under this design at high dimensions. 

Keeping $\ell_{\min}$ fixed while reducing only $\ell_{\max}$ does not scale the entire distribution proportionally. In Appendix~\ref{app:sa-proportional}, we additionally analyze a simple design of proportional scaling of both endpoints, $\ell_{\min} = \alpha \ell_{\max}$ for some fixed $\alpha\in(0, 1)$, and show that it yields a positive limiting expected fraction strictly below one when $\ell_{\max}$ scales with $1/\sqrt{d}$. 

Overall, as $d$ increases, there is a trade-off between a slice's boundary validity and the size of its Euclidean extent, where conditioning on constant scale results in expected validity decaying approximately with $1/d$ and on constant validity results in decaying $\ell$ roughly with $1/\sqrt{d}$. 
Since validity does not necessarily preserve more solver-relevant information, and since in this work we focus on problems of $d\in[2, 10]$ where the constant design gives an expected valid fraction above $0.8$ and does not substantively degenerate in realized rasters, we keep the simple baseline $\ell_{\max} = 1/\sqrt{2}$ as our primary design, with analyses of other values in \S\ref{subsec:rq2-effect_of_probe_and_process}.
\section{Experimental Design}
\label{sec:experimental-design}

\subsection{Problem benchmark and algorithm portfolio}
\label{subsec:benchmark-portfolio}

\paragraph*{Problem benchmark}

We consider the widely used BBOB~\cite{hansen2009real} benchmark as the test suite. For comparability of scale across problems, we evaluate on the 24 one-hot BBOB anchor families instantiated through the MA-BBOB generator~\cite{vermetten2025ma}, which constructs each problem as a weighted geometric aggregation of log-precision-rescaled BBOB base components, parameterized by a weight vector $\vec W$, an instance identifier $\vec I$, and a chosen optimum location $\vec X_{\text{opt}}\in{[-5, 5]^d}$. The problem-wise rescaling factors are derived empirically from taking the median over dimensions on data with 50,000 samples per $(f, d)$ pair, making the problems comparable within orders of magnitude~\cite{vermetten2025ma}. The resulting formulation provides controlled mixtures of BBOB landscapes while retaining the one-hot BBOB cases as interpretable anchors. To instantiate a BBOB instance $(f, d, i)$, we input the one-hot weight vector with $\vec W_f = 1$, instance $\vec I_f = i$, and place the optimum at that of the corresponding BBOB instance. We test across the 24 components, 100 instances each, and dimensions \(d\in\{2,3,5,10\}\), yielding 9,600 instances. 

\paragraph*{Algorithm portfolio}
Our primary portfolio consists of 7 solvers: \texttt{ConfiguredPSO}, \texttt{MultiBFGS}, \texttt{RCobyla}, \texttt{OnePlusOne}, \texttt{DiagonalCMA}, \texttt{modde}, \texttt{moddedefault}, the first 5 of which are accessed through the \texttt{Nevergrad} framework~\cite{bennet2021nevergrad} and the last 2 through modular DE~\cite{vermetten2023modular} (\texttt{modDE}). It is a subset of the nine logged solvers -- which additionally includes \texttt{modcma} and \texttt{modcmaBIPOP} from modular CMA-ES~\cite{van2016evolving} (\texttt{modCMA}) -- that maximizes the empirical VBS-SBS headroom on the frozen $2000d$ AOCC corpus (see Appendix~\ref{subsec:benchmark_analysis}), thereby avoiding a portfolio with little attainable selection gain, following the principle of performance complementarity~\cite{dietrich2024impact}. 

\subsection{Performance targets}
\label{subsec:performance-targets}

We consider two widely used targets, \textit{area over the convergence curve}~\cite{dietrich2024impact} (AOCC) and \textit{relative expected running time}~\cite{prager2022automated} (relERT). Both targets are computed by taking the mean of $R=15$ independent runs of each solver with the common horizon $B=2000d$ evaluations~\cite{dietrich2024impact,vermetten2025ma}. Let $q_{iar}(t)$ denote the best-so-far objective precision of solver $a$ on problem instance $i$ and run $r$ after $t$ evaluations. 

\paragraph*{AOCC}
AOCC reflects how fast a solver reaches high precisions and is a measure of anytime performance of a solver, i.e., evaluating the solver over the entire trajectory of improvement~\cite{hansen2022anytime,lopez2024using}. Following the fixed-target normalization and precision truncation to \([10^{-8},10^2]\) of the empirical attainment curve area~\cite{lopez2024using}, the per-run AOCC is defined as
\[
\mathrm{AOCC}_{iar}
=
\frac{1}{B}\sum_{t=1}^{B}
\frac{2-\log\left(
\operatorname{clip}(q_{iar}(t),10^{-8},10^{2})
\right)}{10},
\]
where the observed best-so-far objective value is carried between logged evaluations. The instance-solver AOCC is 
\begin{equation}\label{eq:aocc}
    \mathrm{AOCC}_{ia}=\frac{1}{R}\sum_r\mathrm{AOCC}_{iar}.
\end{equation} 
AOCC therefore lies in $[0,1]$ and is larger when a solver reaches high precision earlier. 

\paragraph*{relERT}
relERT measures the expected runtime for a solver to reach a prescribed target precision $\tau$, for which we adopt a value widely used in BBO studies, $\tau=10^{-2}$~\cite{prager2022automated,seiler2025deep} (also see Appendix~\ref{subsec:benchmark_analysis}). Let $T_{iar}\leq B$ be the first evaluation for which $q_{iar}(t) \leq\tau$. A successful run contributes to the expected runtime $T_{iar}$ evaluations and an unsuccessful run contributes its observed effort capped at the horizon $E_{iar}=\min(E^{\mathrm{terminal}}_{iar},B)$. Following the COCO convention~\cite{hansen2016coco}, for $s_{ia}>0$ successful runs, the expected running time is
\[
\mathrm{ERT}_{ia}
=
\frac{
\sum_{r:T_{iar}\leq B}T_{iar}
+
\sum_{r:T_{iar}>B}E_{iar}
}{
s_{ia}
},
\label{eq:ert}
\]
which is defined as infinite when there is no successful run. relERT is then given by rescaling by the instance-wise optimum:
\begin{equation}
\mathrm{relERT}_{ia}
=
\frac{\mathrm{ERT}_{ia}}{\min_{b\in\mathcal A:\mathrm{ERT}_{ib}<\infty}}.
\label{eq:relert}
\end{equation}
For undefined instance-solver values,  we adopt the conventional PAR10-style penalty~\cite{prager2022automated,seiler2025deep} and assign them ten times the globally largest finite relERT observed
\begin{equation}
\mathrm{relERT}_{\mathrm{ERT}=\infty}
= 10\cdot \max_{\substack{i,a:\,\mathrm{ERT}_{ia}<\infty}}\mathrm{relERT}_{ia}.
\label{eq:relert-par10}
\end{equation}
Consequently, every retained target row is finite, lower is better, and its
minimum is one.
Instances where \textit{no} solver achieves a finite ERT, i.e. where no solver reaches precision $\tau$ within the evaluation horizon, are omitted, resulting in $8,758$ out of $9,600$ valid cases. This does \textit{not} affect the AOCC instance portfolio.

\subsection{Problem representation}
\label{subsec:representation-panel}

\paragraph*{Probe processor}
We consider two widely used baselines in comparison with $\Phi_{\mathrm{visual}}$, ELA and Deep-ELA~\cite{seiler2025deep}.
Our classical ELA descriptors comprise \texttt{ela\_meta}, \texttt{ela\_distribution}, \texttt{ela\_level}, \texttt{pca}, \texttt{nbc}, \texttt{dispersion}, and \texttt{information\_content}, computed using \texttt{pflacco}~\cite{mersmann2011exploratory,
prager2024pflacco} and orientation-preservingly min-max normalized within each probe. After excluding the extraction-time \texttt{costs\_runtime} outputs, these groups yield 55 descriptors. For Deep-ELA, we adopt the pretrained \texttt{large\_50d\_v1} encoder for the primary $50d$ experiments and \texttt{large\_25d\_v1} at $25d$, both of which output feature vectors of length $48$~\cite{seiler2025deep}. 

\paragraph*{Probe acquisition}
We consider the slice acquisition $\mathcal P_{\mathrm{slice}, b}$ proposed in \S\ref{subsec:geometric-probe} and an additional scrambled Sobol acquisition $\mathcal P_{\text{Sobol}, b}$, each using $b$ objective evaluations per instance. For the slice design, since coordinates outside the problem domain are not evaluated, we keep sampling slices until the total number of evaluations exceeds $b$, in which case the last sampled slice is not evaluated nor used. Here, we adopt the conventional probe budget $b=50d$~\cite{prager2022automated,seiler2025deep} and all slice configurations empirically use resolution $r=5$ (studied in \S\ref{subsec:rq2-effect_of_probe_and_process}). Both designs are deterministic given the seed and nested by budget. Processors compared on a shared acquisition design therefore receive identical objective evaluations, 
as ELA descriptors can be sensitive to the sampling design~\cite{renau2020exploratory}.
The acquisition budget is accounted separately from the $2000d$ solver horizon, with probe evaluations neither deducted from the selected solver's budget nor reused for solving.
Similar to the GeoPAS setting, all probes are min-max normalized instance-wisely before being processed by ELA to improve invariance~\cite{prager2023nullifying}, whereas Deep-ELA encoders have an internal standardization mechanism~\cite{seiler2025deep}, so raw probes are input. 

\subsection{Evaluation pipeline}
\label{subsec:evaluation-pipeline}

\paragraph*{Selector}
The computed representation is then fed into an MLP-based selector $\operatorname{Sel}: H\mapsto \hat a\in\mathcal A$ with two hidden layers of widths $256$ and $128$, ReLU activations, and input dropout $0.2$, which regresses the prescribed targets $\boldsymbol y_{i} \in \mathbb R^{|\mathcal A|}$. Consequently, when the target is AOCC, selection is performed by $\arg\max$ across the predictions, whereas $\arg\min$ is used for relERT. All modules are implemented in \texttt{Python} using \texttt{PyTorch}. For GeoPAS and its variants, the encoder-selector paradigm is trained end-to-end with SmoothL1 loss, whereas only the selector is trained for ELA and Deep-ELA features.

\paragraph*{Validation protocols}



We use two established, complementary protocols, leave-instances-out (LIO) and leave-problem-out (LPO). In LIO, the $100$ instance identifiers are seed-randomly partitioned into five groups, where for each fold all cases carrying the held-out identifiers are tested on the model trained on the rest instances. In LPO, each of the $24$ BBOB problems is held out in turn, with all cases belonging to that problem used for testing. LIO measures transfer to unseen instances of represented problems, and LPO measures transfer to an unseen problem. 
For both protocols, we use three replicates indexed by seeds $16, 17, 18$, independent of the one fixed acquisition seed. In LIO, the replicate seed determines the instance partition. The fold-wise model initialization is deterministically derived from the replicate seed and fold index. Downstream performance is averaged over replicates.

\begin{figure}[t]
    \centering
    \includegraphics[width=\linewidth]
    {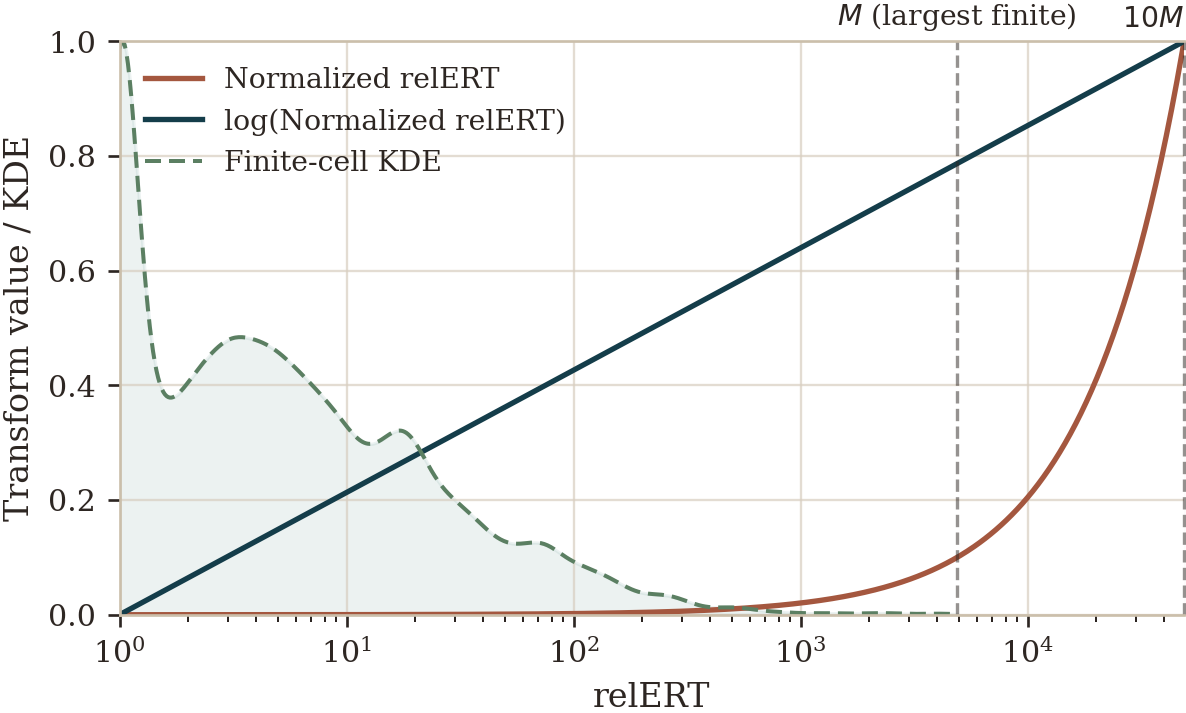}
    \caption{The raw-normalized relERT curve (red) assigns most resolution to the far tail, whereas the logarithmic transformation (navy) distributes resolution over multiplicative performance scales. The green dashed density curve gives a relative KDE of the observed finite relERT values on the log scale.}
    \label{fig:performance-curves}
\end{figure}

\paragraph*{Train-split target transformation}
Since relERT is a ratio to the best algorithm (Equation~\ref{eq:relert}), it is naturally interpreted on a multiplicative scale. 
This is also consistent with the PAR-style treatment of unsuccessful runs, where failures are represented as large, multiplicative runtime penalties.
Indeed, most differences between valid solvers are compressed into a narrow numerical region close to zero (Figure~\ref{fig:performance-curves}). 
This may make catastrophic failures highly visible, but it weakens resolution among ordinary multiplicative performance differences.
To account for that semantic and to limit the influence of the long tail while preserving the within-case ordering, the relERT regression target is logarithmized and then globally min-max normalized as
\begin{equation}
y_{\mathrm{relERT},ia}=
\displaystyle
\frac{\log \mathrm{relERT}_{ia}}{\max_{j\in{\mathcal T},b\in\mathcal A}\log \mathrm{relERT}_{jb}},
\label{eq:relert-transform}
\end{equation}
with $\mathcal T$ the outer training fold and the transformed per-row minimum being $0$. Equal differences in \(\log(\mathrm{relERT})\) correspond to equal multiplicative performance factors in relERT and therefore in ERT. 
For intrinsically normalized AOCC, untransformed $y_{\mathrm{AOCC}, ia} = \mathrm{AOCC}_{ia}$ is used. All reported performance is nevertheless evaluated on the untransformed targets in relERT~\eqref{eq:relert} and AOCC~\eqref{eq:aocc}.

\paragraph*{Reporting metric}
AS, after all, makes sense in principle for its potential to outperform any standalone algorithm. Such a baseline is usually the single best solver (SBS) -- the algorithm achieving the best mean performance on the entire dataset. However, we use the averaged training-fold SBS -- whose global performance is defined as the average over all training-fold SBS's evaluated on the corresponding test splits 
$$
\overline y_{\mathrm{SBS}} = \frac{1}{|F|}\sum_{i=1}^{|F|}\bar f_{\text{train}_i}(\mathcal X_{\text{test}_i})
$$ 
where $F$ is the set of all folds in a protocol and $f_{\text{train}_i} = \arg\operatorname{opt}_{a\in\mathcal A} \bar a(\mathcal X_{\text{train}_i})$ -- as the primary baseline, as we evaluate generalization to held-out within-distribution/instance/problem structure and thereby the static baseline should be estimated under the same information restriction as the selector. 
The virtual best solver (VBS), on the contrary, is the oracle that chooses the best-performing algorithm for each instance and thereby serves as the theoretical limit of any AS system. 
Consequently, SBS is protocol-dependent (e.g. different between LIO and LPO) while VBS is not. 
We primarily report the normalized measure, VBS-SBS gap closure, given by
\begin{equation}
\mathrm{GC}
=
\frac{\overline y_{\mathrm{AS}} - \overline y_{\mathrm{SBS}}}
{\overline y_{\mathrm{VBS}} - \overline y_{\mathrm{SBS}}},
\label{eq:gap-closure}
\end{equation}
with $\vec y$ denoting the target. Thus $0$ is SBS performance, $1$ is VBS performance, and negative values mean that selection is worse than the SBS. 


\subsection{Experimental design}
\label{subsec:rq}

\paragraph*{Questions to investigate} Surrounding the two core ideas of this paper, i.e., slice-based probing and slice-based processing, we design the experiments to clarify the following research questions. 
\textbf{RQ1:} How competitive and complementary is GeoPAS in AS relative to the ELA family? 
\textbf{RQ2:} Which performance differences arise from probe acquisition and which from probe processing? 
\textbf{RQ3:} Which algorithm-performance information and cross-family structure does the learned visual representation contain? 
\textbf{RQ4:} How does GeoPAS performance scale with the probing budget?

For brevity, we let
\begin{equation}
    \mathsf V := \Phi_\text{visual}, \
    \mathsf E := \Phi_\text{ELA}, \
    \mathsf D := \Phi_\text{Deep-ELA}
\end{equation}
denote the visual, ELA, and Deep-ELA processors respectively, and subscripts specify the used probe, e.g. $\mathsf V_{\mathrm{slice}, 50d}$. Brackets $[\cdot]$ denote processors applied to a shared probe, and $\oplus$ denote separately acquired probes. Essentially, $\mathsf V_{\mathrm{slice}, b}$ corresponds to GeoPAS throughout this work. We retain the original notations in RQ2 as the processors and probing strategy are varied internally. 

For RQ1, we first compare the standalone representations under their canonical probe designs, $\mathsf V_{\mathrm{slice},50d}$, $\mathsf E_{\mathrm{Sobol},50d}$, and $\mathsf D_{\mathrm{Sobol},50d}$, under both LIO and LPO on BBOB.  We then evaluate fused representations to test complementarity between them. In particular, fusion is performed by propagating the vector concatenating the involved representations to the selector, trained end-to-end as well. Table~\ref{tab:representation-panel} gives the total-probe-cost-matched $50d$ panel for RQ1.

\begin{table}[t]
\centering
\small
\caption{Representation panel for RQ1.}
\label{tab:representation-panel}
\begin{tabular}{@{}p{0.24\linewidth}p{0.7\linewidth}@{}}
\toprule
Feature group& Configurations \\
\midrule
Standalone& $\mathsf V_{\mathrm{slice}, 50d}$, $\mathsf E_{\mathrm{Sobol}, 50d}$, $\mathsf D_{\mathrm{Sobol}, 50d}$\\
\midrule
Shared-probe fusion& $[\mathsf V, \mathsf E]_{\mathrm{slice}, 50d}$, $[\mathsf V, \mathsf D]_{\mathrm{slice}, 50d}$, $[\mathsf V, \mathsf E, \mathsf D]_{\mathrm{slice}, 50d}$, $[\mathsf E, \mathsf D]_{\mathrm{Sobol}, 50d}$\\
\midrule
Separate probe fusion& 
$\mathsf V_{\mathrm{slice}, 25d} \oplus  \mathsf E_{\mathrm{Sobol}, 25d}$, $\mathsf V_{\mathrm{slice}, 25d} \oplus \mathsf D_{\mathrm{Sobol}, 25d}$, $\mathsf V_{\mathrm{slice}, 25d} \oplus [\mathsf E, \mathsf D]_{\mathrm{Sobol}, 25d}$\\
\bottomrule
\end{tabular}
\end{table}

For RQ2, we separately vary how the probes are acquired and the samples are processed, with the other kept fixed. For probing effect, we evaluate all standalone processors across different configurations of $\mathcal P_{\mathrm{slice}}$, with variations in the slice scale distribution parameters $(\ell_{\min}, \ell_{\max})$ and the grid resolution $r$. For processing effect, we test different choices for several modules of $\Phi_{\mathrm{visual}}$, with different probe components, slice-based-CNN $\phi_{\mathrm{slice}}$ replaced with slice-based ELA, disabling per-slice normalization as well as the range/IQR conditioning, and disabling the dimension conditioning.  

For RQ3, we analyze frozen out-of-fold representations from $\mathsf V_{\mathrm{slice},50d}$, $\mathsf E_{\mathrm{Sobol},50d}$, and $\mathsf D_{\mathrm{Sobol},50d}$ using the same low-capacity linear probe. First, we measure how directly each representation exposes per-algorithm performance information through target prediction and the resulting selection performance. Second, we test incremental predictability after conditioning on another representation family, asking whether the visual representation contains solver-relevant information not already accessible from ELA or Deep-ELA. Finally, we compare the representation spaces using a common similarity measure, providing a representation-level view of their overlap and complementarity.


For RQ4, we evaluate $\mathsf V_{\mathrm{slice},b}$ and $\mathsf E_{\mathrm{Sobol},b}$ over the nested nominal budgets $b/d\in\{50,100,150,200,300\}$.


\paragraph*{Other implementation details}
Experiments are conducted on a dual-socket server equipped with two AMD EPYC 7763 CPUs and eight NVIDIA RTX 3090 GPUs, running \texttt{Ubuntu 22.04.5 LTS}. Each per-fold model is trained with optimizer \texttt{Adam}, learning rate $10^{-3}$, fixed epoch number $100$, batch size $64$, and weight decay $3\times10^{-4}$. Owing to the modest model size, each per-fold model is trained on a single GPU, and training one model takes approximately 4 minutes.
\section{Results}
\label{sec:results}

\begin{figure*}[t]
    \centering
    \includegraphics[width=\linewidth]{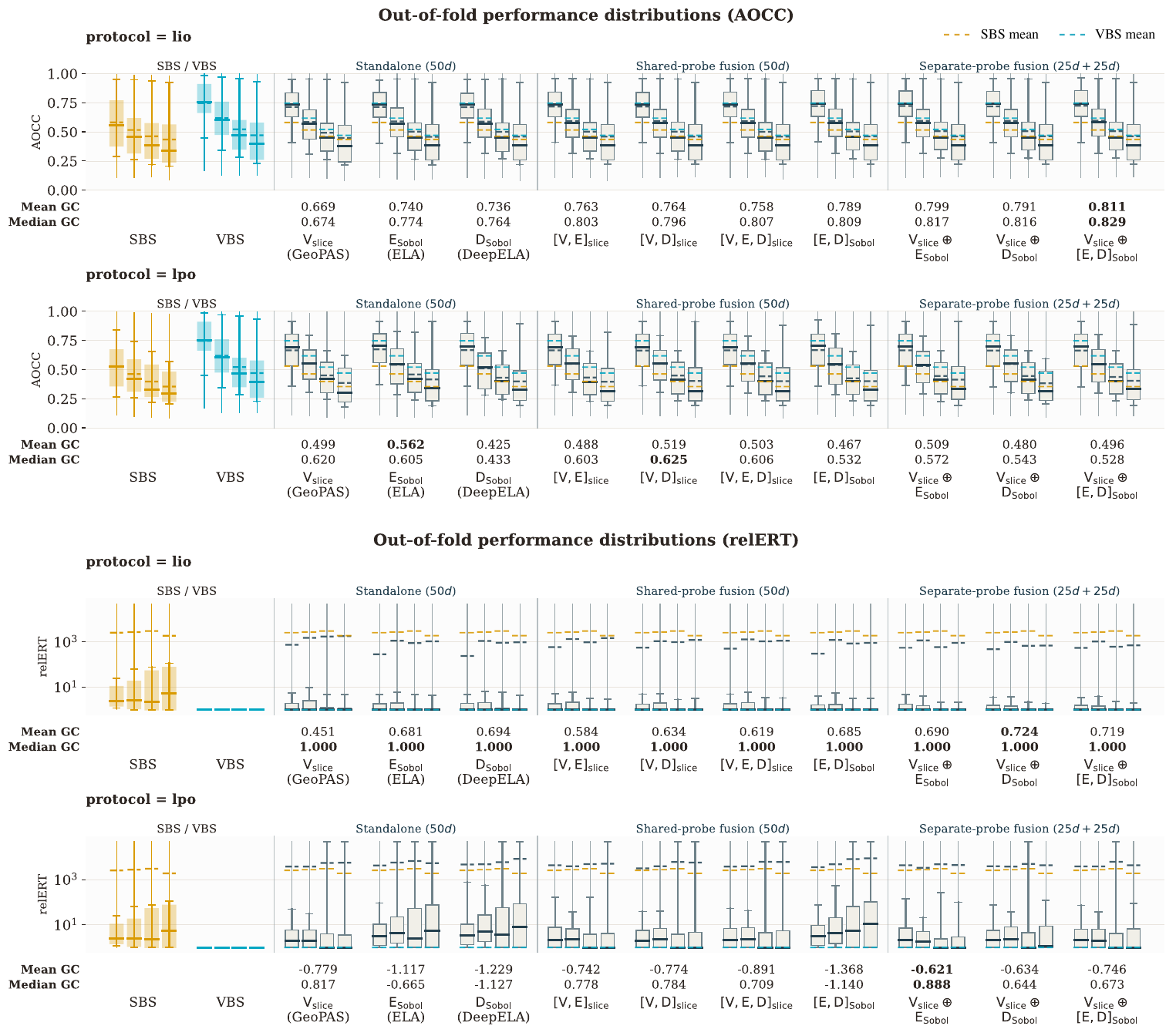}
    \caption{Out-of-fold achieved AOCC/relERT \emph{seed–case-level} outcome-distribution intervals, including the training-fold SBS and case-wise VBS references. Every method position contains four adjacent box-and-whisker glyphs, ordered left to right by $d\in\{2,3,5,10\}$. Boxes span Q25–Q75; solid and dashed bars show the median and arithmetic mean respectively; and capped whiskers span Q10–Q90 and thin whiskers the full range. Short amber and cyan dashed ticks superposed on each learned-selector glyph mark the corresponding dimension-specific SBS and VBS means respectively. relERT-axis is in log-scale. The mean/median gap-closure rows retain the \emph{pooled} values over the seeds.}
    \label{fig:main_results}
\end{figure*}

\subsection{\textbf{RQ1:} Competitiveness and complementarity}
Figure~\ref{fig:main_results} compares the standalone representations and their cost-matched fusions. Under LIO, every configuration closes a positive fraction of the VBS--SBS gap in both mean and median for both targets. For relERT, all median GCs equal one, meaning that a case-wise best solver is selected in at least half of the pooled seed--case outcomes. This also makes the median non-discriminative in this setting, leaving the mean and upper distribution to distinguish the methods.

The standalone group differs qualitatively between interpolation and problem-level transfer. Under LIO, $\mathsf V_{\mathrm{slice}}$ is behind both ELA-family representations, with its mean GC $0.669$ for AOCC and $0.451$ for relERT compared with $0.740/0.681$ for $\mathsf E_{\mathrm{Sobol}}$ and $0.736/0.694$ for $\mathsf D_{\mathrm{Sobol}}$. On these within-problem splits, GeoPAS is therefore not as effective as the established descriptors when used alone. Under LPO, this ordering changes. For AOCC, $\mathsf V$ obtains mean/median GC $0.499/0.620$, where its mean remains below $E$ at $0.562$ but its median is slightly higher than $\mathsf E$'s $0.605$, and both values exceed $\mathsf D$'s $0.425/0.433$. For relERT, $V$ reaches $-0.779/0.817$, compared with $-1.117/-0.665$ for $\mathsf E$ and $-1.229/-1.127$ for $\mathsf D$. Its mean is still below the SBS, but its typical transfer is materially stronger.

Looking into dimension-wise distributions, AOCC decreases with $d$ for the SBS and VBS as well as for every learned selector, supporting that the absolute decline is primarily a property of the benchmark regime. Under relERT--LIO, the SBS deteriorates with dimension while the central selector distributions all remain close to the case-wise optimum. Under relERT--LPO, the $\mathsf V$-free representations instead retain an SBS-like dimension pattern and increasingly long upper ranges at $d=5$ and $10$. Representations containing $V$ lower the central distributions across dimensions, most visibly at the two higher dimensions, but do not remove the extreme upper tail. This is the distributional counterpart of their positive median GC and negative mean GC.

Fusion is most consistently beneficial under LIO. For AOCC, every fused representation exceeds the best standalone method in both mean and median GC. The strongest result is $\mathsf V_{\mathrm{slice},25d}\oplus[\mathsf E,\mathsf D]_{\mathrm{Sobol},25d}$, with mean/median GC $0.811/0.829$. Under relERT--LIO, where all medians are already saturated at one, $\mathsf V_{\mathrm{slice},25d}\oplus \mathsf D_{\mathrm{Sobol},25d}$ gives the highest mean GC, $0.724$. Moreover, each separate-probe fusion has higher mean GC than its shared-slice counterpart under both LIO targets, despite the same total $50d$ evaluation budget.

It however does not seem to be a generic consequence of concatenating more features. Among the shared-probe variants, $[\mathsf V,\mathsf E,\mathsf D]_{\mathrm{slice}}$ is never the best configuration. Within the ELA family, $[\mathsf E,\mathsf D]_{\mathrm{Sobol}}$ improves AOCC--LIO, is approximately neutral relative to the standalone ELA families under relERT--LIO, and deteriorates markedly under relERT--LPO. The separate-probe advantage is not uniform under LPO either. Standalone $\mathsf E_{\mathrm{Sobol}}$ gives the best AOCC mean GC, $0.562$, while $[\mathsf V,\mathsf D]_{\mathrm{slice}}$ gives the best median, $0.625$. In contrast, $\mathsf V_{\mathrm{slice},25d}\oplus \mathsf E_{\mathrm{Sobol},25d}$ is the strongest relERT--LPO configuration in both metrics, $-0.621/0.888$.

Looking into the selection diagnostics (Appendix~\ref{app:rq1_selection_behavior}), $\mathsf V$-containing selectors do not obtain their positive median GC by reverting to the SBS, but depart from it more often and agree more frequently with the VBS, while the remaining harmful switches account for their negative mean GC.

Overall, the results support complementarity between the visual and ELA-family representations, but only conditional on how the probe information is acquired and processed. The particularly consistent LIO advantage of separate-probe fusion suggests that complementary acquisition together with a processor specialized to each probe may matter more than feature count alone. This figure does not isolate those two factors, which are then examined in RQ2.

\subsection{\textbf{RQ2:} Acquisition versus processing} \label{subsec:rq2-effect_of_probe_and_process}

Tables~\ref{tab:effect-probing-strategy} and~\ref{tab:effect-probe-processing} separate the contributions from what is sampled and how it is encoded by varying the acquisition and processing stages in turn.
\begin{table*}[t]
\centering
\caption{Acquisition effects with the processing architecture fixed. Entries are pooled mean GC. The rows of each panel -- resolution or scale distribution -- vary one component of the baseline slice design. $\Phi_\mathrm{visual}$ is not compatible with a Sobol probe so the corresponding observations are not available. Within each panel, each column-wise best entry is in \textbf{bold}.}
\label{tab:effect-probing-strategy}
\resizebox{\linewidth}{!}{%
\begin{tabular}{l|ccc|ccc|ccc|ccc}
\toprule
\textbf{Probe}& \multicolumn{6}{c|}{$\mathrm{AOCC}$ \textbf{mean GC}}
& \multicolumn{6}{c}{$\mathrm{relERT}$ \textbf{mean GC}}\\
\cmidrule{2-7}\cmidrule{8-13}& \multicolumn{3}{c|}{\textbf{LIO}}
& \multicolumn{3}{c|}{\textbf{LPO}}
& \multicolumn{3}{c|}{\textbf{LIO}}
& \multicolumn{3}{c}{\textbf{LPO}}\\
\cmidrule{2-4}\cmidrule{5-7}\cmidrule{8-10}\cmidrule{11-13}& $\Phi_{\mathrm{visual}}$& $\Phi_{\mathrm{ELA}}$& $\Phi_{\mathrm{DeepELA}}$& $\Phi_{\mathrm{visual}}$& $\Phi_{\mathrm{ELA}}$& $\Phi_{\mathrm{DeepELA}}$& $\Phi_{\mathrm{visual}}$& $\Phi_{\mathrm{ELA}}$& $\Phi_{\mathrm{DeepELA}}$& $\Phi_{\mathrm{visual}}$& $\Phi_{\mathrm{ELA}}$& $\Phi_{\mathrm{DeepELA}}$\\
\midrule
 \multicolumn{13}{l}{\textbf{Resolution $r$ (baseline: $r=5$)}}\\
 $r = 1$ ($\mathcal P_{\mathrm{Sobol}}$) & ---& \textbf{0.740}& \textbf{0.736}& ---& \textbf{0.562}& 0.425& ---& 0.681& \textbf{0.694}& ---& -1.117&-1.229\\
 $r=3$& \textbf{0.679}& 0.702& 0.729& 0.458& 0.554& 0.46& \textbf{0.548}& \textbf{0.684}& 0.651& \textbf{-0.575}& -0.999&-1.091\\
 $r = 5$ (baseline) & 0.669& 0.649& 0.62& 0.499& 0.538& 0.534& 0.451& 0.512& 0.441& -0.779& -1.078&-1.257\\
$r=7$& 0.633& 0.619& 0.562& \textbf{0.512}& 0.518& \textbf{0.575}& 0.448& 0.354& 0.288& -0.738& \textbf{-0.99}& -1.077\\
$r=10$& 0.563& 0.555& 0.499& 0.481& 0.521& 0.551& 0.407& 0.166& 0.276& -0.841& -1.168& \textbf{-0.891}\\
\midrule
 \multicolumn{13}{l}{\textbf{Scale $\ell$ distribution (baseline: $\log \ell \sim \mathcal{U}(\log \ell_{\min}, \log \ell_{\max}), \ell_{\min} = 0.02, \ell_{\max} = 1/\sqrt{2}$)}}\\
$\ell_{\max} = 0.5$& 0.667& 0.653& 0.616& 0.499& \textbf{0.56}& \textbf{0.543}& 0.497& 0.492& 0.456& -0.769& -1.162& -1.291\\
$\ell_{\max} = 1/\sqrt{2}$ (baseline) & 0.669& 0.649& 0.62& 0.499& 0.538& 0.534& 0.451& 0.512& 0.441& -0.779& -1.078&-1.257\\
$\ell_{\max} = 1$& 0.663& 0.649& 0.624& \textbf{0.507}& 0.547& 0.541& 0.479& 0.544& 0.482& \textbf{-0.731}& -1.078& \textbf{-1.151}\\
$\ell_{\max} = 1/\sqrt{d}$& \textbf{0.679}& 0.65& 0.615& 0.486& 0.54& 0.524& 0.435& 0.462& 0.42& -0.775& -1.263& -1.269\\
$\ell \sim \mathcal{U}(\ell_{\min}, \ell_{\max})$& 0.673& \textbf{0.67}& \textbf{0.691}& 0.501& 0.538& 0.507& \textbf{0.543}& \textbf{0.648}& \textbf{0.572}& -0.775& \textbf{-0.882}& -1.292\\
\bottomrule
\end{tabular}%
}
\end{table*}

Changing ELA and Deep-ELA from their canonical Sobol probes to the slice probe reduces LIO GC in both metrics, whereas $\Phi_{\mathrm{visual}}$ leads both processors for AOCC and is comparable with Deep-ELA for relERT. Under LPO, the Sobol--slice differences are mixed and change sign. This becomes more interesting when considered together with the resolution changes. Indeed, note that since the slice centers follow a Sobol design, $\mathcal P_{\mathrm{slice}}$ with $r = 1$ essentially \emph{produces a Sobol set of evaluations} equivalent to $\mathcal P_{\mathrm{Sobol}}$, which can be viewed as least local concentration and most spread, while higher resolution can be viewed as concentrating more at fewer local regions. With that relation, we see that the performances of all processors nearly monotonically deteriorate as $r$ increases, with the only exception being a small improvement of ELA for relERT from $r = 1$ to $r=3$. Under LPO, by contrast, the optima vary across processors and targets, with a rough advantaged interval around $r=7$ for GeoPAS and Deep-ELA. Generally, a coarser raster permits more restrictions within the budget, and the within-distribution result consistently favors breadth across slices over density, whereas this trade-off does not transfer uniformly under problem transfer. In comparison, scale has a milder effect on AOCC. For LIO, linear-uniform scale sampling gives the highest GC for all cases except for LIO--$\Phi_{\mathrm{visual}}$ whose performance is nevertheless comparable to the best case. The dimension-adaptive upper scale $\ell_{\max}=1/\sqrt d$, despite mitigating dimension-dependent boundary loss, yields no systematic selection gain. 
Boundary validity alone is therefore insufficient for choosing the scale distribution.

\begin{table}[t]
\centering
\caption{Processing effects on the shared baseline $50d$ slice probe. Entries are pooled mean GC over three training seeds; higher is better. The first block changes the processor, and the second ablates components of $\Phi_{\mathrm{visual}}$. Incomplete configurations are omitted.}
\label{tab:effect-probe-processing}
\resizebox{\linewidth}{!}{%
\begin{tabular}{l|cc|cc}
\toprule
\textbf{Processor} &
\multicolumn{2}{c|}{$\mathrm{AOCC}$ \textbf{mean GC}} &
\multicolumn{2}{c}{$\mathrm{relERT}$ \textbf{mean GC}}\\
\cmidrule{2-3}\cmidrule{4-5}
& \textbf{LIO} & \textbf{LPO} & \textbf{LIO} & \textbf{LPO}\\
\midrule
$\Phi_{\mathrm{visual}}$ & 0.669 & 0.499 & 0.451 & -0.779\\
$\Phi_{\mathrm{ELA}}$    & 0.649 & 0.538 & 0.512 & -1.078\\
$\Phi_{\mathrm{DeepELA}}$& 0.620 & 0.534 & 0.441 & -1.257\\
\midrule
\multicolumn{5}{l}{\textbf{Global probe normalization}}\\
No global normalization        & 0.748& 0.436& 0.695&-1.106\\
\midrule
\multicolumn{5}{l}{\textbf{Visual-processor ablations}}\\
No per-slice normalization        & 0.717& 0.43& 0.7&-1.006\\
No amplitude statistics $(\Delta_i,q_i)$ & 0.657& 0.532& 0.398&-0.833\\
No scale information $\ell_i$    & 0.682& 0.486& 0.466&-0.758\\
No dimension information $d$     & 0.728& 0.492& 0.502&-0.681\\
\midrule
\multicolumn{5}{l}{\textbf{Target transformation}}\\
Raw& ---& ---& -1.373& -1.848\\
Train-fold normalized & 0.703& 0.488& 0.432&-1.007\\
\bottomrule
\end{tabular} %
}
\end{table}

Holding the slice probe fixed exposes a corresponding processing effect. $\Phi_{\mathrm{visual}}$ is strongest for AOCC--LIO and, more distinctly, for relERT--LPO; ELA or Deep-ELA leads under AOCC--LPO and ELA under relERT--LIO. In particular, GeoPAS's relERT--LPO advantage persists when all processors receive the same observations, locating that advantage in the visual processing pipeline rather than in slice acquisition alone.

Global probe normalization gives the clearest architectural trade-off. Removing it raises LIO GC by $0.079$ for AOCC and $0.244$ for relERT, while reducing the corresponding LPO values by $0.063$ and $0.327$. The invariance to probe-level objective offset and amplitude therefore discards useful within-distribution cues, yet materially improves transfer across held-out functions. The remaining conditioning variables contribute less consistently. Removing scale information changes GC by at most $0.021$, while removing dimension conditioning improves three of the four mean GCs, including relERT--LPO by $0.098$. These results do not support dimension conditioning as necessary in the present configuration.

The two controls collectively show that the RQ1 differences belong to acquisition--processor pairings. Sobol probing particularly benefits the ELA families within distribution, while the visual processor better preserves relERT selection quality under problem-level shift. This division is also consistent with the advantage of separate-probe fusion in RQ1, which preserves each processor's specialized acquisition.

\subsection{\textbf{RQ3:} Information contained in the learned representation}\label{subsec:rq3}
To examine the information exposed by the representations independently of the nonlinear selector, we use the same outer folds and standalone $50d$ representations as in RQ1. Each outer-fold GeoPAS encoder is frozen, and ridge probes of the same specification are fitted separately to its outer-training representations, to the corresponding $\mathsf E_{\mathrm{Sobol},50d}$ and $\mathsf D_{\mathrm{Sobol},50d}$ descriptors. Evaluation uses the matched held-out cases. A dimension-only probe is retained as a control. Training targets follow the transformation described in \S\ref{subsec:evaluation-pipeline}. We report held-out \textit{squared-error skill}, i.e., out-of-sample squared-error performance relative to the per-solver training-mean predictor, together with the resulting GC. The probe definition is given in Appendix~\ref{app:rq3}.

\begin{figure}[t]
    \centering
    \includegraphics[width=\linewidth]{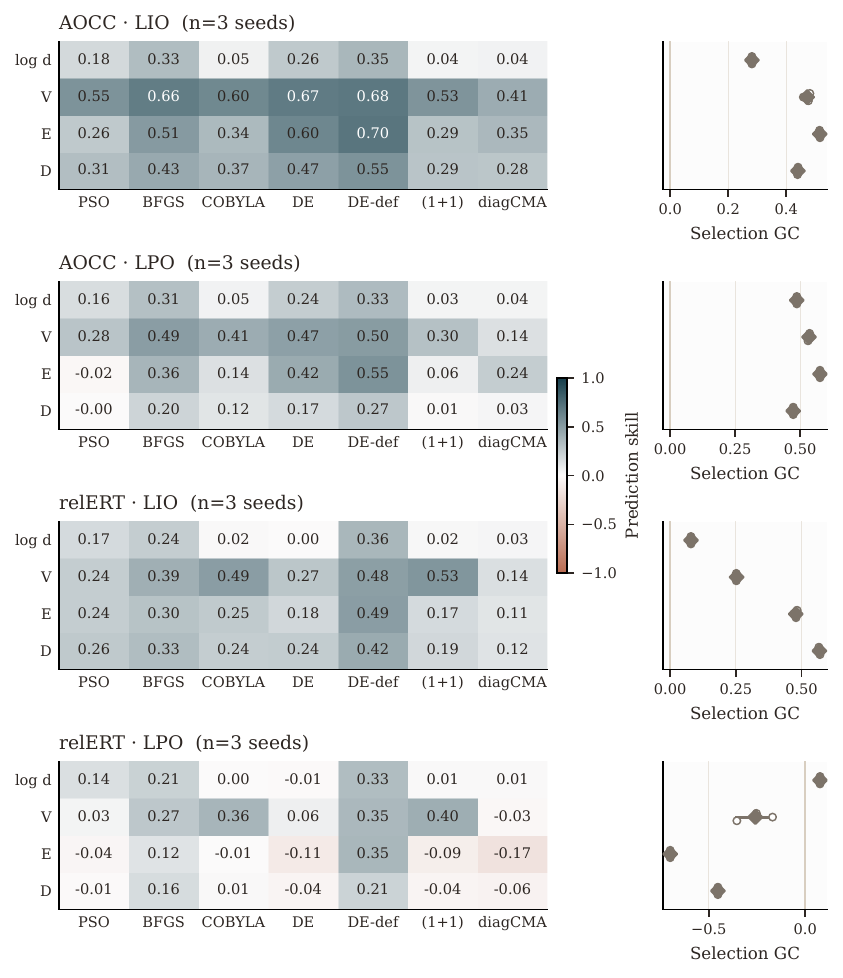}
    \caption{Accessibility of solver performance from the frozen representations. Heatmaps report held-out prediction skill for each solver, averaged over three training seeds, and the aligned panels report the GC obtained by selecting from the same predictions. Diamonds denote seed means, circles individual seeds, and horizontal segments their ranges.}
    \label{fig:rq3-accessibility}
\end{figure}

Figure~\ref{fig:rq3-accessibility} first shows that $\mathsf V$ gives the highest macro skill among the three standalone representations in every target--protocol setting and the highest per-solver skill in $22$ of the $28$ corresponding comparisons. This advantage however does not translate monotonically into selection performance. $\mathsf V$ is not the best standalone selector in either AOCC setting or under relERT--LIO; under relERT--LPO, it performs best among $\mathsf V$, $\mathsf E$, and $\mathsf D$, but still remains below the training-fold SBS. Indeed, the dimension-only probe attains the only marginally positive GC, $0.077$. 

\begin{figure}[t]
    \centering
    \includegraphics[width=\linewidth]{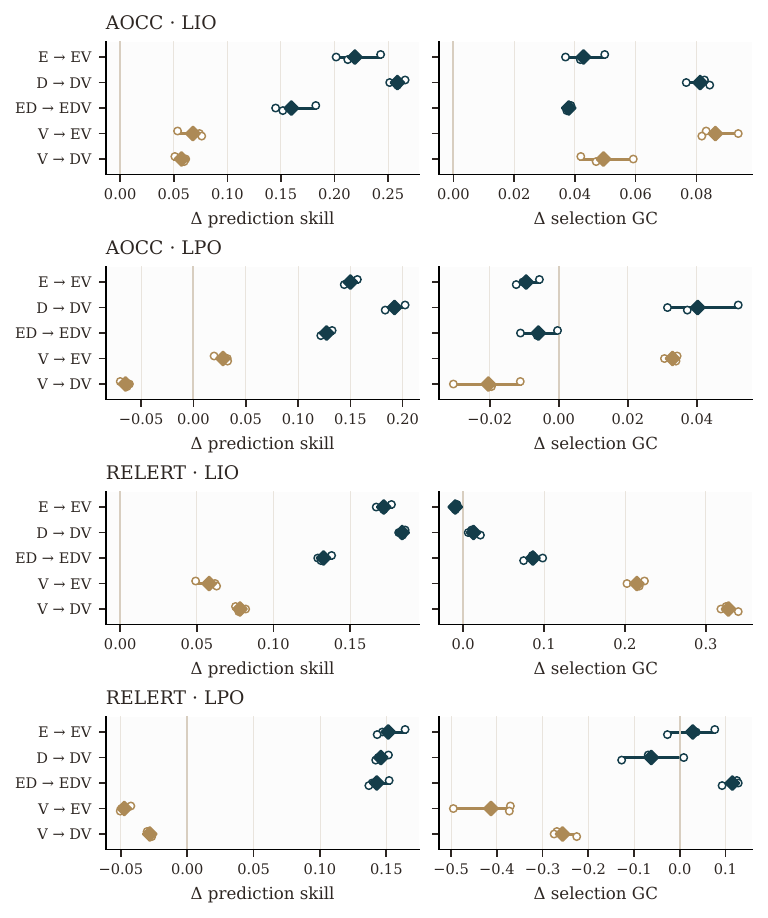}
    \caption{Incremental accessibility from augmenting one representation with another. The left panels report changes in macro prediction skill and the right panels changes in selection GC; positive values favor the augmented representation. $\mathsf E\rightarrow \mathsf E\oplus \mathsf V$, $\mathsf D\rightarrow \mathsf D\oplus \mathsf V$, and $[\mathsf E,\mathsf D]\rightarrow [\mathsf E,\mathsf D]\oplus \mathsf V$ test what $\mathsf V$ adds to the ELA families, while the last two rows give the reverse comparisons. Since $\mathsf V_{\mathrm{slice}}$ and $\mathsf E_{\mathrm{Sobol}}$/$\mathsf D_{\mathrm{Sobol}}$ use separate $50d$ probes, these combinations have total acquisition cost $100d$ and are therefore not budget-matched comparisons.}
    \label{fig:rq3-incremental}
\end{figure}

The incremental comparisons in Figure~\ref{fig:rq3-incremental} give the more direct evidence of complementarity. Adding $\mathsf V$ to $\mathsf E$, $\mathsf D$, or $[\mathsf E,\mathsf D]$ increases absolute-target prediction skill in every target--protocol setting and every seed. For the most comprehensive baseline, $[\mathsf E,\mathsf D]\rightarrow [\mathsf E,\mathsf D]\oplus \mathsf V$, the mean increases range from $0.127$ to $0.160$ across the settings, whereas the corresponding GC changes are from $-0.006$ to $0.115$. The predictive increment is therefore consistent, whereas its decision value is conditional. The reverse additions are likewise not uniformly beneficial. Under relERT--LPO, for example, adding either $\mathsf E$ or $\mathsf D$ to $\mathsf V$ reduces both prediction skill and GC.

Absolute performance prediction may recover common problem difficulty without equally recovering the relative advantages that determine selection. We therefore repeat the skill calculation for the within-case centered target vector. Under AOCC--LPO, $\mathsf V$ obtains contrast skill $0.071$, compared with $0.086$ for the dimension-only control, despite its substantially stronger absolute-target skill. In the other three settings, $\mathsf V$ remains the strongest standalone representation on the contrast task. These scores use different reference errors and should not be read as a decomposition of the absolute-target score; the per-solver results are reported in Appendix Figure~\ref{fig:rq3-contrast}.

We also perform a simpler fold-local linear CKA, showing moderate overall alignment between $\mathsf V$ and the ELA families, but substantially lower alignment after broad group structure is removed. Under LPO, for example, $\mathsf V$--$\mathsf E$ and $\mathsf V$--$\mathsf D$ CKA decrease from approximately $0.49$ overall to $0.14$--$0.17$ after dimension-mean removal; $\mathsf E$--$\mathsf D$ decreases from $0.69$--$0.71$ to $0.29$--$0.30$. The complete comparison is in Appendix Figure~\ref{fig:rq3-cka}. CKA here contextualizes the predictive results; lower alignment alone is not evidence of useful complementarity.

Overall, the learned visual representation exposes solver-performance information directly accessible to a linear decoder and incrementally accessible beyond ELA and Deep-ELA. The same results also delimit this conclusion: some of that information concerns absolute difficulty rather than relative solver advantage, and increased accessibility does not by itself ensure mean-safe selection under problem-level shift.

\subsection{\textbf{RQ4:} Probe-budget efficiency}

\begin{figure}
    \centering
    \includegraphics[width=\linewidth]{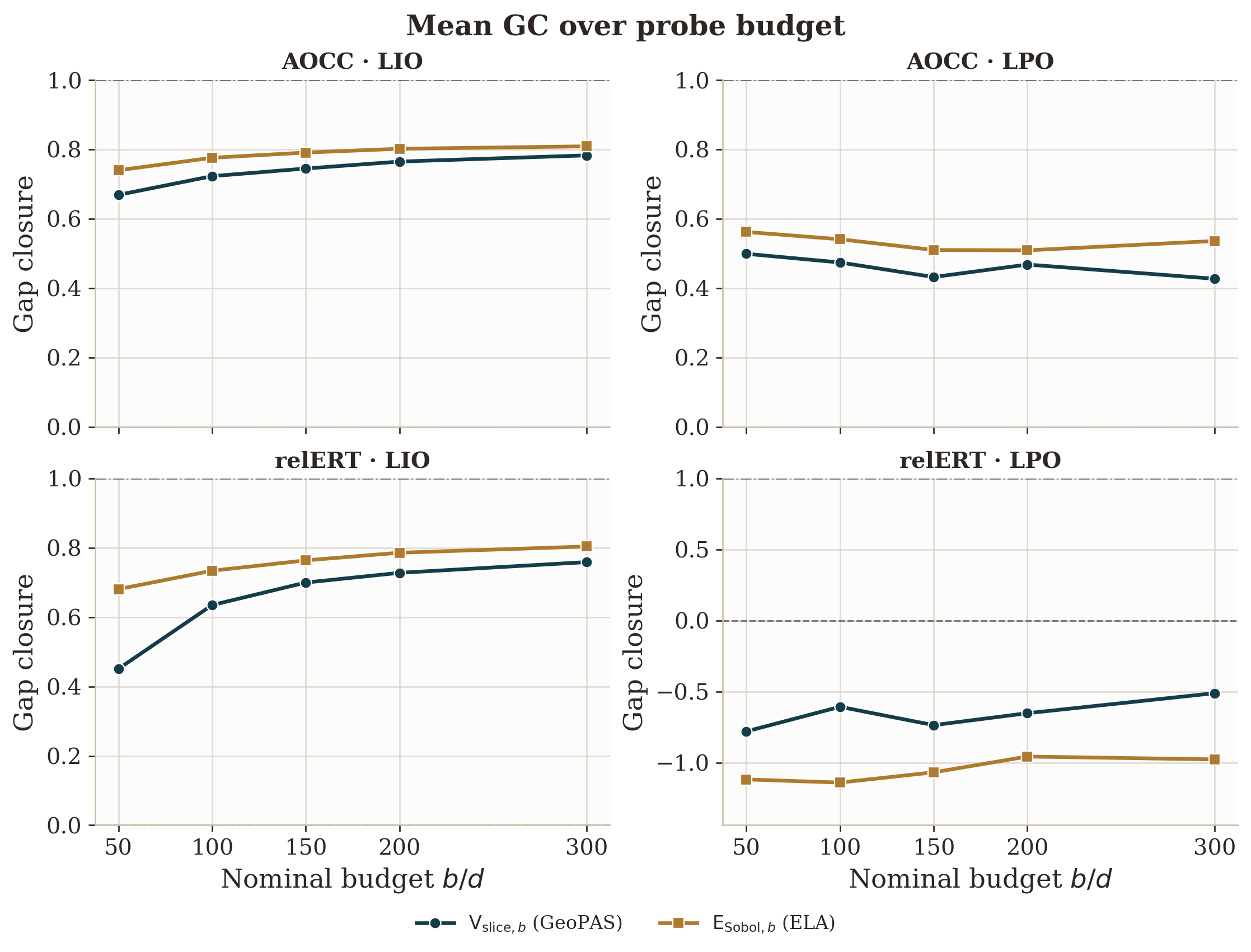}
    \caption{Mean GC of GeoPAS and ELA over nested nominal probe budgets $b/d\in\{50,100,150,200,300\}$. Curves average three training seeds. Horizontal references denote the VBS ($\mathrm{GC}=1$) and SBS ($\mathrm{GC}=0$). The SBS reference coincides with the lower boundary of the nonnegative panels.}
    \label{fig:probe-budget}
\end{figure}

Figure~\ref{fig:probe-budget} extends the canonical $50d$ comparison using nested probes. Under LIO, both representations improve with budget and exhibit diminishing returns. 
GeoPAS rises from $\sim0.669$ to $\sim0.78$ on AOCC and from $\sim0.451$ to $\sim0.76$ on relERT; ELA rises from $\sim0.740$ to $\sim0.81$ and from $\sim0.681$ to $\sim0.80$, respectively. 
ELA remains stronger at every shown budget, while its lead contracts from $\sim0.071$ to $\sim0.03$ for AOCC and from $\sim0.230$ to $\sim0.05$ for relERT. 
The visual representation is thus more constrained by the $50d$ probe, particularly for relERT, and receives most of its additional benefit by $100d$--$150d$.

The LPO response is qualitatively different. For AOCC, neither curve improves over its $50d$ value, and both vary non-monotonically as observations are added. For relERT, GeoPAS improves overall from $-0.779$ to approximately $-0.51$ and remains above ELA throughout; ELA changes from $-1.117$ to approximately $-0.98$. Both remain below the SBS at every budget, with intermediate reversals along each curve. Additional probe information therefore narrows some deficits but does not produce reliable problem-level transfer.

The budget curves separate two limitations. At LIO, the low-budget GeoPAS deficit is substantially reduced by additional slice coverage. At LPO, information volume alone is insufficient: the remaining failure concerns how the acquired structure generalizes to unseen function families and how its errors translate into solver decisions. The reported mean curves establish this average budget response; seed-level stability requires the corresponding dispersion statistics.

\section{Discussion}
\label{sec:discussion}
\paragraph*{Implications for geometric probing}
GeoPAS contributes a distinct acquisition--processing bias, not a generally stronger descriptor. Its relative advantage under problem-level shift persists for relERT when all processors receive the same slice observations, locating that advantage within the visual processing itself. The normalization ablation further identifies invariance to global objective offset and scale as one contributor: it reduces discrimination within represented functions while improving robustness across functions. The fusion results then acquire a sharper meaning. Under LIO, combining normalized local geometry with global Sobol descriptors is most effective when each processor retains its associated acquisition; further encodings of one shared probe offer no comparable benefit. The operative complementarity lies in what is observed and what is made invariant, not in feature count.

The resolution, scale, and budget experiments jointly indicate what the slice probe should preserve. Within distribution, allocating a fixed budget across more coarse restrictions is preferable to resolving fewer restrictions densely, and most benefit from additional evaluations is obtained by approximately $100d$--$150d$. Yet the dimension-adaptive scale limit improves boundary validity without improving selection. The relevant quantity is therefore the diversity of solver-discriminative local geometries, for which neither pixel density nor in-domain coverage is an adequate surrogate. Under LPO, the non-monotone budget response further rules out insufficient observation count as the sole limitation. Together with the normalization and dimension-conditioning ablations, it places the remaining difficulty in the stability of the learned representation--performance relationship; explicit dimension information may even facilitate a benchmark-level shortcut.

The representation analysis makes this limitation precise. GeoPAS adds linearly accessible performance variation beyond ELA and Deep-ELA, but absolute performance accessibility can largely reflect common problem difficulty, whereas selection depends on within-instance solver contrasts. For relERT, optimizing prediction on the logarithmic scale also gives limited control over rare choices that dominate mean native runtime. Hence additional features or evaluations need not improve GC, and can be harmful under shift, as the dimension-only control illustrates. Further improvement should focus on decision-aligned learning---within-case ranking, selection regret, tail-sensitive objectives, and calibrated fallback---rather than continued enlargement of the representation. Since GeoPAS is performance-supervised while ELA and Deep-ELA are fixed descriptors, this conclusion concerns information made accessible by the trained systems, not an intrinsic ordering of their underlying probes.

\paragraph*{Future work}
On acquisition, the slice construction is not intrinsically dimension-limited, so a natural next step is to evaluate beyond $d\leq 10$ while varying scale and boundary treatment jointly. Their appropriate relation depends on the domain semantics. Under our assumption of the box being a hard constraint, feasibility geometry is part of the problem, while BBOB problems are not intrinsically constrained, and $[-5, 5]^d$ only defines an area of interest. Metric-aware or feasibility-aware restrictions may then be more meaningful than maximizing in-domain coverage alone.

On processing, GeoPAS deliberately omits the sampled locations $X$ and the positional parameters $(c,O)$, thereby removing direct access to ambient pose while retaining scale. Controlled reintroduction of this information, through low-capacity conditioning or equivariant processing, would establish when this bias removes nuisance variation and when it suppresses solver-relevant structure. The same acquisition–processing comparison should also be extended to MA-BBOB~\cite{vermetten2025ma} and real-world-informed benchmarks~\cite{rodriguez2025mechbench,ivic2025randomness}, using suite-aware portfolios and protocols that separate within-corpus generalization from cross-suite transfer.


Other interesting directions include generalizing the probe acquisition beyond isotropic boxes via metric-aware scaling and feasibility-geometry conditioning, and extending the GeoPAS pipeline to multi-objective BBO problems, which would then require a permutation- and scale-aware representation of vector-valued slice observations. 


\section*{Acknowledgment}
The authors used ChatGPT to assist with language polishing and manuscript organization. 
All scientific claims, experiments, analyses, and text were finalized, reviewed, and approved by the authors, who take full responsibility for the content.



\bibliographystyle{IEEEtran}
\bibliography{references}

\clearpage

{\appendices

\appendix

\subsection{Slice analytics supplements}\label{app:slice_analytics_supplements}

\subsubsection{Exact expectation proof}
\label{app:sa-exact}
The product in~\eqref{eq:sa-product} expands as
\begin{equation}
    \prod_{i=1}^d(1-\sigma|V_i|)
    =\sum_{I\subseteq\{1,\ldots,d\}}
    (-\sigma)^{|I|}\prod_{i\in I}|V_i|.
\end{equation}
Observe that coordinates of $V$ are exchangeable as permuting them still results in a $\mathcal U(\mathbb S^{d-1})$ variable. Therefore every subset $I$ of the same size $m$ has the same \textit{moment}. Since for each $m\in\{1,...,d\}$, there are $d\choose{m}$ subsets, we have
\begin{equation}\label{eq:sa-exchangeable-expansion}
    p_d(\sigma) = \sum_{m=0}^d (-\sigma)^m {\binom dm}\mathbb E\left[\prod_{i=1}^m|V_i|\right],
\end{equation}
where the empty product is one.

We next compute these moments. Write $V=Z/\|Z\|_2$, where the $Z_i$ are independent standard normal variables. Then $Z_i^2/2$ are independent Gamma variables with shape $1/2$ and unit scale. Their normalized values satisfy
\begin{equation}
    (V_1^2,\ldots,V_d^2)
    \sim\operatorname{Dirichlet}(1/2,\ldots,1/2).
\end{equation}
For a Dirichlet vector $W$ with parameters $\alpha_i$, $\alpha_0=\sum_i\alpha_i$ and powers $\gamma_i\ge0$, integration of its density gives
\begin{equation}
    \mathbb E\left[\prod_{i=1}^d W_i^{\gamma_i}\right]
    =\frac{\Gamma(\alpha_0)}
    {\Gamma(\alpha_0+\sum_i\gamma_i)}
    \prod_{i=1}^d\frac{\Gamma(\alpha_i+\gamma_i)}
    {\Gamma(\alpha_i)}.
\end{equation}
Indeed, multiplication by the monomial changes each exponent of the density from $\alpha_i-1$ to $\alpha_i+\gamma_i-1$, so the integral is the ratio of the two normalizing constants. Applying this identity with $W_i=V_i^2$ and $\gamma_i=1/2$ for $i\le m$, and zero otherwise, gives
\begin{equation}\label{eq:sa-sphere-moment}
    \mathbb E\left[\prod_{i=1}^m|V_i|\right]
    =\frac{\Gamma(d/2)}
    {\pi^{m/2}\Gamma((d+m)/2)},\qquad 0\le m\le d.
\end{equation}
Substitution into~\eqref{eq:sa-exchangeable-expansion} results in
\begin{equation}\label{eq:sa-exact-radial}
    p_d(\sigma)=\sum_{m=0}^d(-\sigma)^m\binom dm
    \frac{\Gamma(d/2)}{\pi^{m/2}\Gamma((d+m)/2)}.
\end{equation}
Thus the expectation is given by
$$
\begin{aligned}
    &\bar\eta_d(\ell_{\min}, \ell_{\max})  = \mathbb E_{\sigma}[p_d(\sigma)] \\
    =& \sum_{m=0}^d(-1)^m\binom dm
    \frac{\Gamma(d/2)}{\pi^{m/2}\Gamma((d+m)/2)} \cdot \mathbb E[\sigma^m] \\
    = &\sum_{m=0}^d(-1)^m\binom dm
    \frac{\Gamma(d/2)}{\pi^{m/2}\Gamma((d+m)/2)} \cdot \mathbb E\left[(L\|U\|_2)^m\right],
\end{aligned}
$$
where independence of $L$ and $U$ implies
\begin{equation}
    \mathbb E[(L\|U\|_2)^m]
    =\mathbb E[L^m]\mathbb E[\|U\|_2^m].
\end{equation}
For $m\geq1$, the log-uniform density gives
\begin{equation}
\begin{aligned}
    \mathbb E[L^m] &= \frac1{\log(\ell_{\max}/\ell_{\min})}
    \int_{\ell_{\min}}^{\ell_{\max}}\ell^{m-1}\,d\ell \\
    &= \frac{\ell_{\max}^m-\ell_{\min}^m}
    {m\log(\ell_{\max}/\ell_{\min})}.
\end{aligned}
\end{equation}
For $\mathbb E[\|U\|_2^m]$, observe that the square has eight congruent polar sectors. Without loss of generality, on the sector $0\le\theta\le\pi/4$, its radial boundary is $r=(2\cos\theta)^{-1}$, thus we have
$$
\begin{aligned}
    E[\|U\|_2^m]&=8\int_0^{\pi/4}\int_0^{(2\cos\theta)^{-1}}
    r^{m+1}\,dr\,d\theta\\
    &=\frac{2^{1-m}}{m+2}\int_0^{\pi/4}
    \sec^{m+2}\theta\,d\theta.
\end{aligned}
$$
These integer moments have elementary expressions. In particular, if
\begin{equation}
    J_n:=\int_0^{\pi/4}\sec^n\theta\,d\theta,
\end{equation}
integration by parts gives
\begin{equation}
    J_n=\frac{2^{(n-2)/2}}{n-1}
    +\frac{n-2}{n-1}J_{n-2},\qquad n\ge3,
\end{equation}
with $J_1=\log(1+\sqrt2)$ and $J_2=1$. 

\subsubsection{Dimension scaling}\label{app:sa-scaling}
We first show the $\|V\|_1$ limit. With the aforementioned normalized Gaussian representation, as $d$ increases, we have
\begin{equation}
    \frac{\|V\|_1}{\sqrt d}
    =\frac{d^{-1}\sum_{i=1}^d|Z_i|}
    {\sqrt{d^{-1}\sum_{i=1}^d Z_i^2}}
    \xrightarrow{\mathbb P}\sqrt{\frac2\pi},
\end{equation}
by the law of large numbers. Consistently,~\eqref{eq:sa-sphere-moment} with $m=1$ gives
\begin{equation}
    \mathbb E\|V\|_1
    =\frac{d\,\Gamma(d/2)}{\sqrt\pi\,\Gamma((d+1)/2)}
    \sim\sqrt{\frac{2d}\pi}.
\end{equation}

For the probability limit, we fix $t\ge0$, let $\sigma_d=t/\sqrt d$, and define
\begin{equation}
    q_d(V):=\prod_{i=1}^d\max(1-\sigma_d|V_i|, 0).
\end{equation}
For sufficiently large $d$, $\sigma_d<1$ and the $\max$ part can be omitted. The bound
\begin{equation}
    0\le-\log(1-x)-x\le\frac{x^2}{2(1-x)},
    \qquad 0\le x<1,
\end{equation}
follows by summing the power series of $-\log(1-x)$. Since $\sum_i V_i^2=1$, it implies
\begin{equation}\label{eq:sa-log-bound}
    0\le-\log q_d(V)-\sigma_d\|V\|_1
    \le\frac{\sigma_d^2}{2(1-\sigma_d)}\longrightarrow0.
\end{equation}
Since $\sigma_d\|V\|_1\xrightarrow{\mathbb P} t\sqrt{2/\pi}$, we have $q_d(V)\xrightarrow{\mathbb P}\exp{\left(-t\sqrt{2/\pi}\right)}$. As $q_d$ is bounded between zero and one, its expectation converges to the same constant. 

For a fixed $\sigma>0$, the inequality $\max(1-x, 0)\le e^{-x}$ gives
\begin{equation}\label{eq:sa-product-upper}
    0\le\prod_i\max(1-\sigma|V_i|, 0)
    \le e^{-\sigma\|V\|_1}\xrightarrow{\mathbb P}0.
\end{equation}
Boundedness again gives $p_d(\sigma)\to0$. When $L$ has a fixed positive scale interval, $L\|U\|_2>0$ almost surely. Dominated convergence in~\eqref{eq:sa-radial-average} therefore gives $\bar\eta_d\to0$.

\subsubsection{Proportional scaling of the scale interval}
\label{app:sa-proportional}
The main-text comparison fixes $\ell_{\min}=0.02$ and reduces only $\ell_{\max}$. This interval narrows as $d$ increases and is nondegenerate only for $d<2500$. To study a distribution that can be scaled consistently across dimensions, we instead fix $0<\alpha<1$ and $\kappa>0$ and let
\begin{equation}\label{eq:sa-proportional-endpoints}
    \ell_{\min}(d)=\alpha\kappa d^{-\beta},
    \qquad\ell_{\max}(d)=\kappa d^{-\beta},
\end{equation}
where $\beta\in\mathbb R$. Equivalently, $L_d=\kappa d^{-\beta}T$, with $\log T\sim\mathcal U(\log\alpha,0)$. 

\begin{proposition}[Proportional scale design]
\label{prop:sa-proportional}
Under~\eqref{eq:sa-proportional-endpoints},
\begin{equation}\label{eq:sa-three-regimes}
    \bar\eta_d\longrightarrow
    \begin{cases}
        0,&\beta<1/2,\\
        F_\alpha(\kappa),&\beta=1/2,\\
        1,&\beta>1/2,
    \end{cases}
\end{equation}
where
\begin{align}
    F_\alpha(\kappa)
    &:=\mathbb E_{T,U}\left[e^{-\mu\kappa T\|U\|_2}\right]
    \label{eq:sa-limit-coverage}\\
    &=\frac1{\log(1/\alpha)}
    \int_\alpha^1\int_S e^{-\mu\kappa t\|u\|_2}\,du\,\frac{dt}{t}
    \in(0,1).
\end{align}
\end{proposition}
\begin{IEEEproof}
Fix $t\in[\alpha,1]$ and $u\ne0$, and set $\sigma_d=\kappa t\|u\|_2d^{-\beta}$. Then
\begin{equation}
    A_d:=\sigma_d\|V\|_1
    =\kappa t\|u\|_2d^{1/2-\beta}
    \frac{\|V\|_1}{\sqrt d}.
\end{equation}
If $\beta<1/2$, then $A_d\to\infty$ in probability, and the upper bound in~\eqref{eq:sa-product-upper} gives $q_d(V)\to0$ in probability. If $\beta>1/2$, then $A_d\to0$ in probability and
\begin{equation}
    0\le1-q_d(V)
    \le\sum_i\min\{\sigma_d|V_i|,1\}\le A_d,
\end{equation}
so $q_d(V)\to1$ in probability. For $\beta=1/2$,~\eqref{eq:sa-critical-limit} applies with $t$ there replaced by $\kappa t\|u\|_2$. In each case, boundedness gives convergence of the expectation over $V$. Dominated convergence then permits averaging over $T,U$. The excluded event $U=0$ has probability zero. Finally, at $\beta=1/2$ the limiting integrand is positive and strictly below one almost everywhere, proving $0<F_\alpha(\kappa)<1$.
\end{IEEEproof}

Thus $\beta=1/2$ is the unique exponent in this proportional family yielding a limiting expected fraction strictly between zero and one. In particular, $\ell_{\max}=1/\sqrt d$ and $\ell_{\min}=\alpha\ell_{\max}$ give the limit $F_\alpha(1)$. For $\alpha=0.01$, numerical integration gives $F_{0.01}(1)\approx0.9397$. A fraction close to one therefore need not converge to one. The result is asymptotic and does not assert exact constancy, or monotonicity, at finite $d$.

Figure~\ref{fig:sa-proportional} illustrates the proportional-lower-endpoint design. For the adaptive maximum, the displayed expected fraction levels off as $d$ increases. With the fixed maximum, $\ell_{\min}=\alpha/\sqrt2$ is also fixed, and the expected continuous fraction instead tends to zero. The realized raster distributions provide a separate description of how validity varies between sampled slices.

\begin{figure}[t]
    \centering
    \includegraphics[width=0.75\linewidth]
    {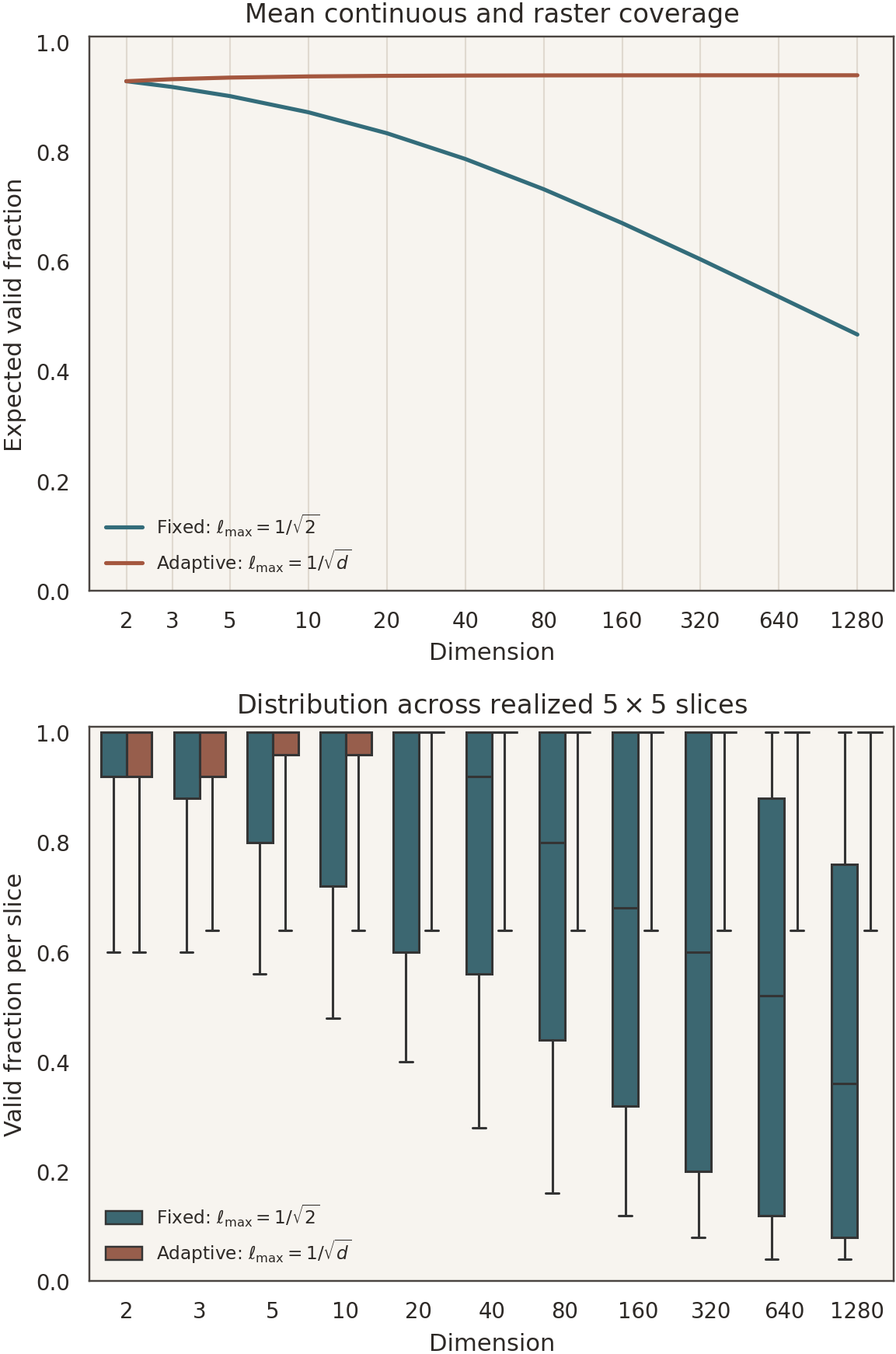}
    \caption{Boundary validity, under the two scale designs, with $\ell_{\min}=\alpha\ell_{\max}$ for a fixed $\alpha\in(0,1)$. The two designs use $\ell_{\max}=1/\sqrt2$ and $1/\sqrt d$. (Top) the expected continuous valid fraction and (bottom) the distribution of realized $5\times5$ raster fractions. The adaptive design approaches $F_\alpha(1)$.}
    \label{fig:sa-proportional}
\end{figure}

For fixed $\alpha$, $F_\alpha(\kappa)$ is continuous and strictly decreasing in $\kappa\ge0$, with $F_\alpha(0)=1$ and $F_\alpha(\kappa)\to0$ as $\kappa\to\infty$. These properties follow directly from the integrand and dominated convergence. Consequently, each prescribed limiting fraction in $(0,1)$ determines a unique $\kappa>0$. This permits geometric calibration once the desired fraction and $\alpha$ have been chosen; it does not determine those choices.

At the critical exponent, the Euclidean displacement satisfies
\begin{equation}
    \|L_d\mathcal O u\|_2
    \le\frac{\kappa}{\sqrt{2d}}.
\end{equation}
Maintaining a positive limiting fraction therefore comes with shrinking spatial extent. We include this design as an extension of the geometric analysis; whether its validity calibration is useful for downstream algorithm selection remains a question for future work.

\subsection{Benchmark analysis}\label{subsec:benchmark_analysis}

\begin{figure}[t]
    \centering
    \includegraphics[width=\linewidth]{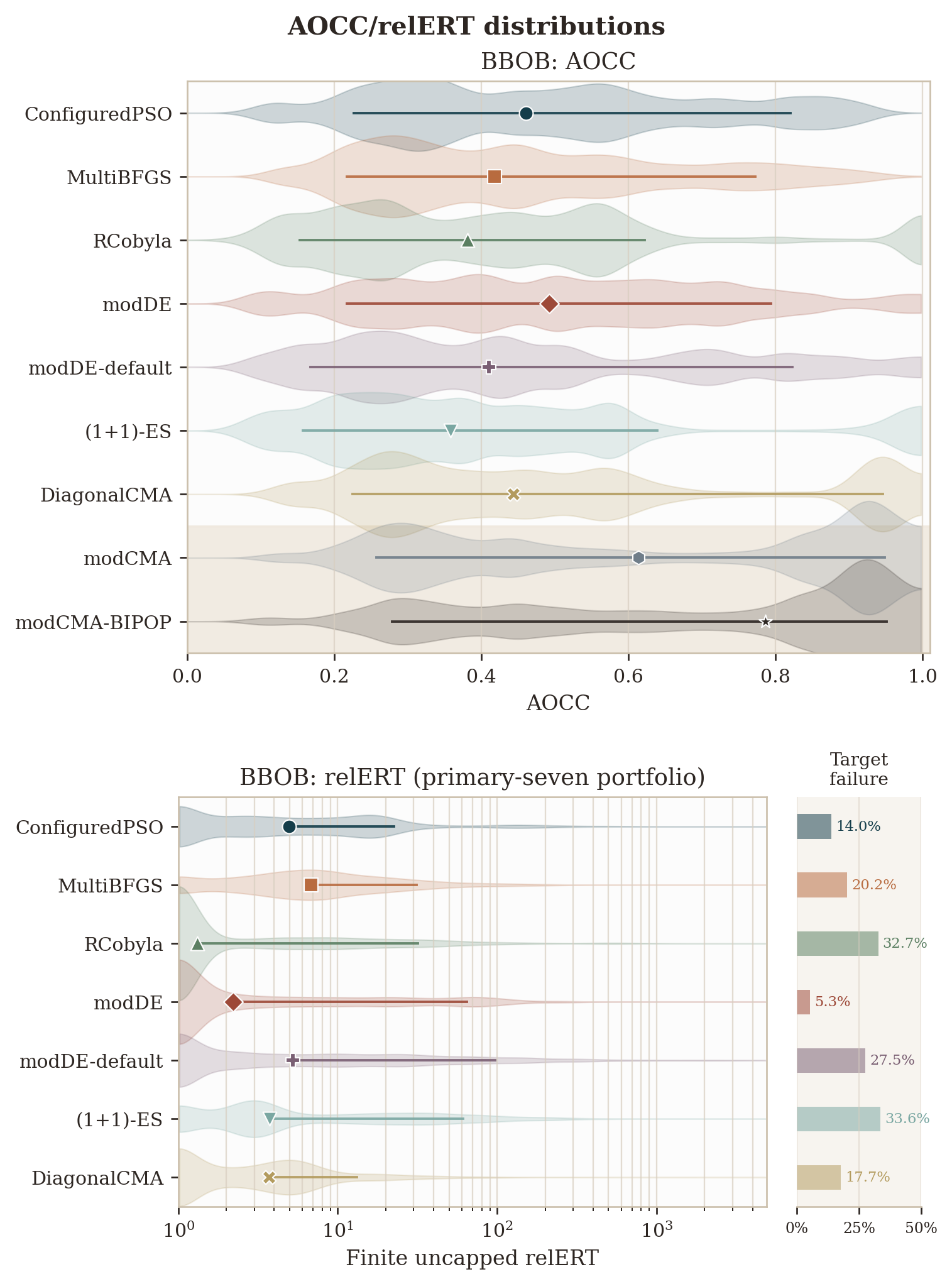}
    \caption{Solver performances at budget $2000d$ and target precision $10^{-2}$ for 9,600 BBOB cases. AOCC includes all nine methods, while relERT are computed with the seven max-headroom methods. relERT ridges show only uncapped values, with the adjacent bar indicating the proportions of failures. Ridges have equal area in the plotted coordinate (linear AOCC; $\log_{10}$ relERT), and share a vertical scale within each metric, Smoothing is boundary-corrected at both AOCC limits and at the relERT lower limit. Points mark medians; segments span P10–P90 for AOCC and the median–P90 interval for finite relERT. Shading identifies the two additional modCMA variants in the AOCC panels.}
    \label{fig:benchmark_profiles}
\end{figure}


\paragraph*{Portfolio complementarity}
A useful algorithm-selection benchmark requires scope for instance-dependent choices to improve on a fixed solver~\cite{dietrich2024impact}. We therefore choose the primary portfolio by maximizing the empirical AOCC headroom on the frozen $2000d$ corpus: the difference between the mean instance-wise best performance and the best mean performance of a single solver. Figure~\ref{fig:benchmark_profiles} shows the strong performance of the two additional modular CMA-ES variants. The seven-solver portfolio retains more variation in the leading solver, although \texttt{DiagonalCMA} still leads many groups. The headroom criterion additionally accounts for their magnitudes. The resulting selection task, and hence the conclusions drawn from it, are conditional on this portfolio.


\begin{figure*}
    \centering
    \includegraphics[width=0.8\linewidth]{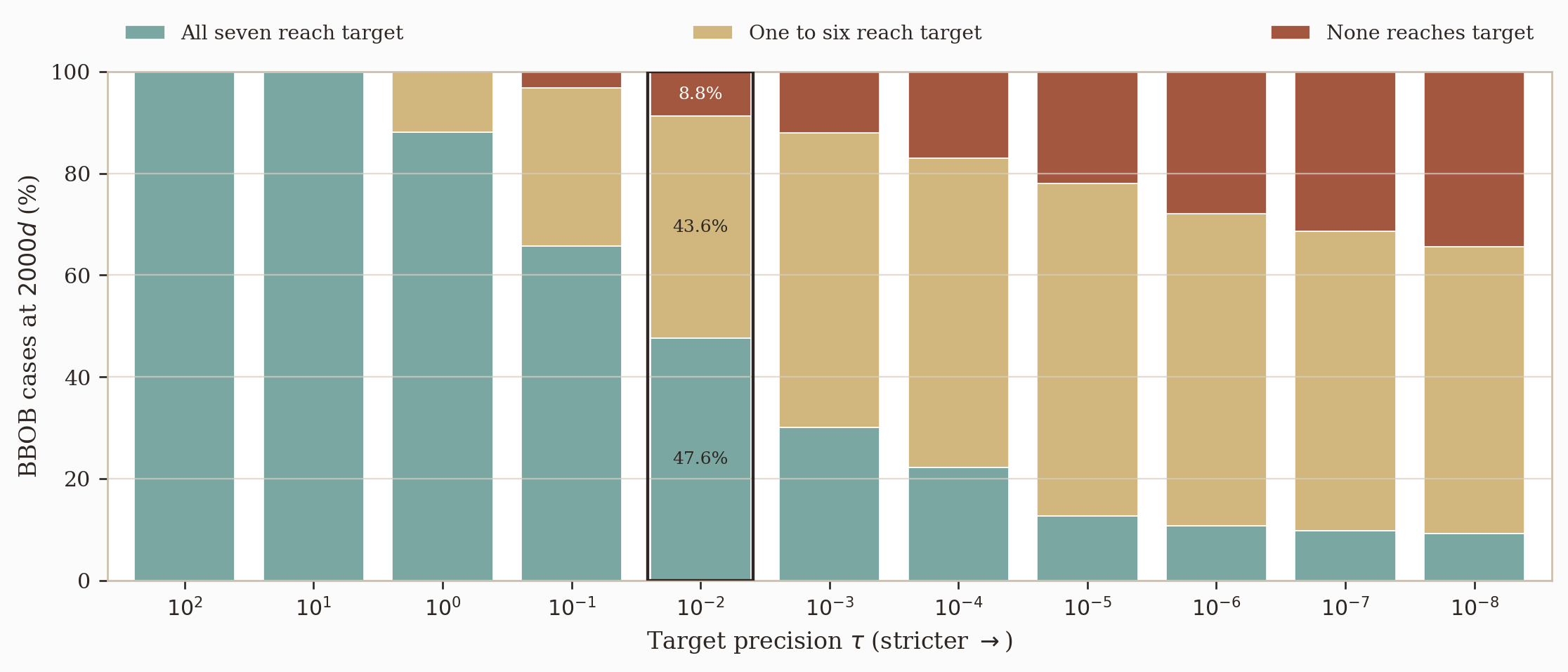}
    \caption{Seven-method BBOB target reachability across precision. Each bar partitions all 9,600 BBOB cases according to how many portfolio members reach target precision $\tau$ within $2000d$: all seven, between one and six, or none. The outlined bar marks the selected $\tau=10^{-2}$.}
    \label{fig:benchmark_reachability}
\end{figure*}

\paragraph*{Target precision}
The target $\tau=10^{-2}$ follows previous AS studies~\cite{prager2022automated,seiler2025deep} and provides a practical balance between reachability and discrimination within $2000d$ evaluations. At this precision, all seven solvers reach the target on $47.6\%$ of BBOB cases, only a subset does so on $43.6\%$, and none does so on $8.8\%$ (Figure~\ref{fig:benchmark_reachability}). Solver choice can therefore affect both runtime and whether the target is reached, while $91.2\%$ of cases remain evaluable by relERT. Looser targets reduce the distinction in reachability; stricter targets progressively increase the fraction of cases excluded from relERT. This supports the chosen operating point without establishing it as uniquely preferable.

\begin{figure}[t]
    \centering
    \includegraphics[width=\linewidth]{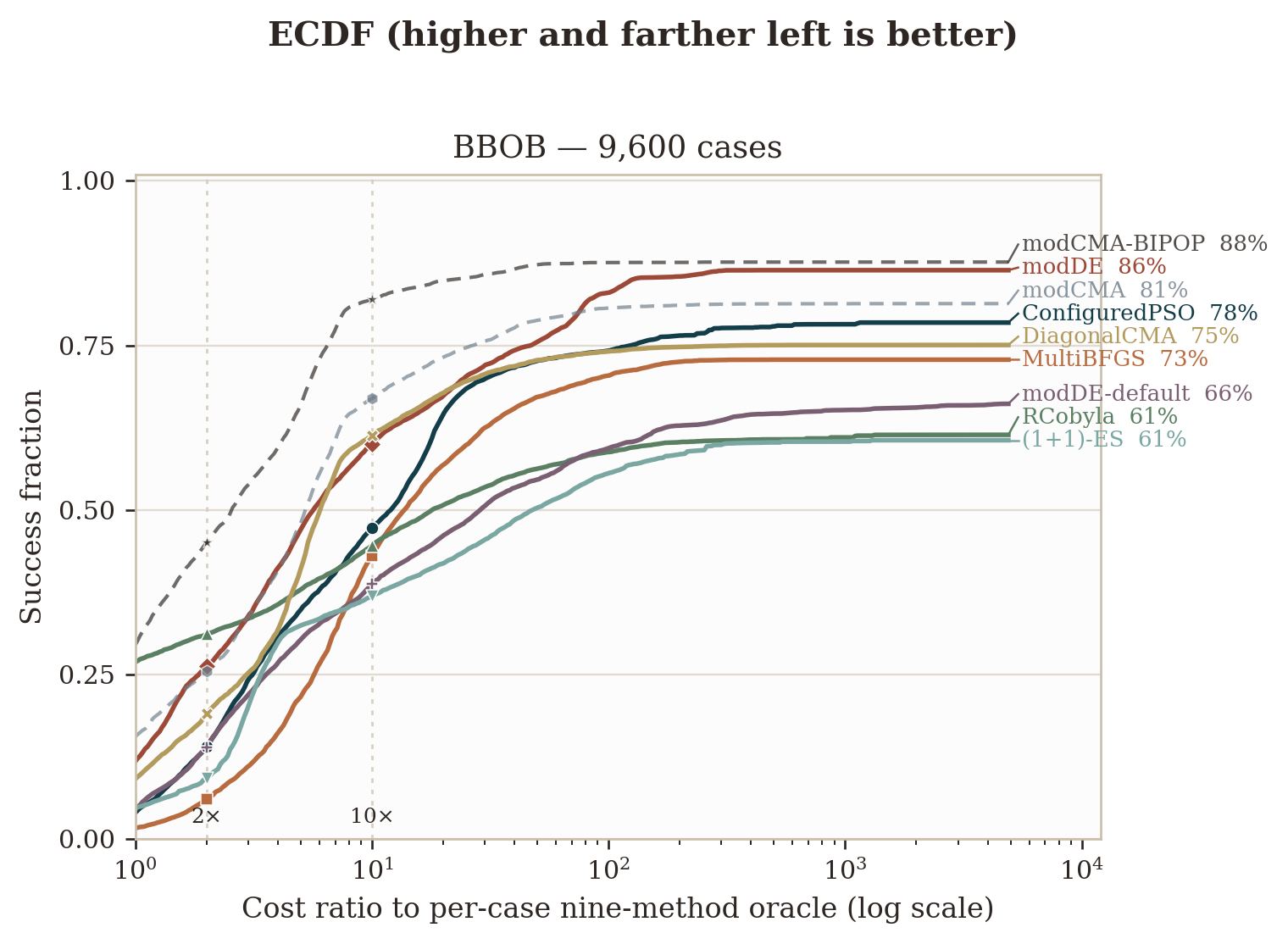}
    \caption{Empirical cumulative target-hitting profiles.}
    \label{fig:benchmark_ecdf}
\end{figure}

\begin{figure*}
    \centering
    \includegraphics[width=0.7\linewidth]{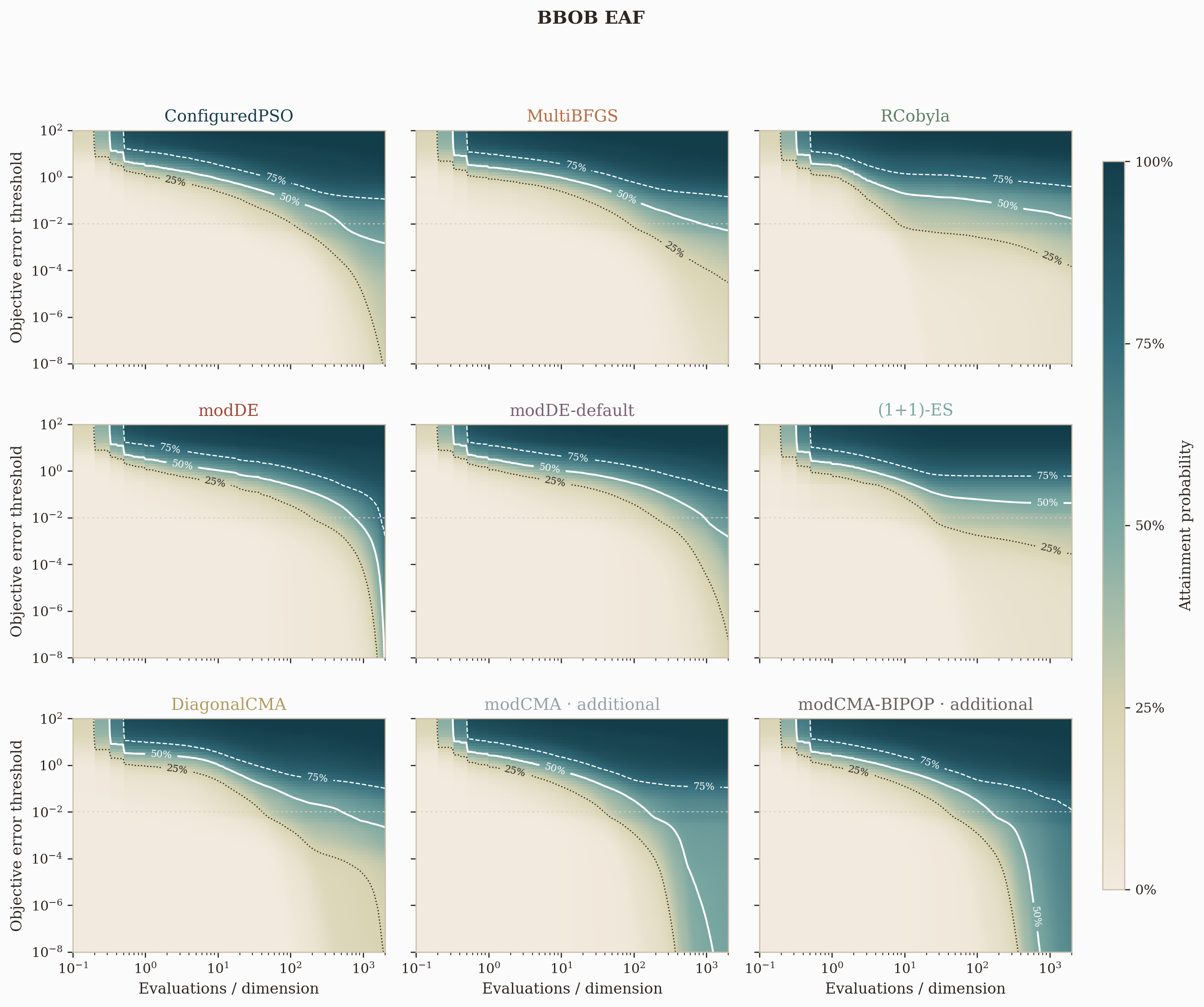}
    \caption{Method-specific empirical attainment functions. Colour gives $\alpha_a(t,\varepsilon)$ over 144,000 equally weighted run–case pairs per method (9,600 cases $\times$ 15 runs), with evaluations normalized by dimension. Dark dotted, white solid, and white dashed curves mark the 25\%, 50\%, and 75\% attainment boundaries, respectively. The horizontal reference marks objective error $10^{-2}$.}
    \label{fig:benchmark_eaf_bbob}
\end{figure*}


\paragraph*{Runtime and anytime performance}
Finite relERT distributions must be read together with target failures (Figure~\ref{fig:benchmark_profiles}): \texttt{RCobyla}, for example, combines competitive runtimes on successfully solved cases with substantial failure. The empirical cumulative distribution functions (ECDFs) retain this distinction through their horizontal positions and terminal heights (Figure~\ref{fig:benchmark_ecdf}). The attainment surfaces show the underlying dependence on both evaluation budget and target precision (Figures~\ref{fig:benchmark_eaf_bbob}). We therefore retain AOCC alongside relERT to assess progress over the optimization trajectory, including cases where the prescribed target remains unreached.

\begin{table*}[t]
    \centering
    \caption{Performance and selection-behavior statistics for the RQ1 representation panel. Entries report mean $\pm$ standard deviation over three training seeds. Conditional switch gain is reported in the native target scale and is therefore not comparable in magnitude between AOCC and relERT.}
    \begin{tabular}{ll|cccc}
        \toprule
        \textbf{Feature group} & \textbf{Feature}
            & \textbf{Selection entropy}
            & \textbf{SBS agreement}
            & \textbf{VBS agreement}
            & \textbf{Conditional switch gain} \\
        \midrule
    
        \multicolumn{6}{l}{\textbf{AOCC--LIO}}\\
        Standalone
            & $\mathsf V_{\mathrm{slice},50d}$
            & $0.902 \pm 0.007$
            & $0.164 \pm 0.012$ & $0.507 \pm 0.016$ & $0.072 \pm 0.003$\\
            & $\mathsf E_{\mathrm{Sobol},50d}$
            & $0.900 \pm 0.002$
            & $0.242 \pm 0.013$ & $0.546 \pm 0.013$ & $0.088 \pm 0.002$\\
            & $\mathsf D_{\mathrm{Sobol},50d}$
            & $0.900 \pm 0.007$
            & $0.238 \pm 0.015$ & $0.565 \pm 0.009$ & $0.087 \pm 0.001$\\
        \cmidrule(lr){1-6}
        Shared probe
            & $[\mathsf V,\mathsf E]_{\mathrm{slice},50d}$
            & $0.916 \pm 0.003$
            & $0.239 \pm 0.010$ & $0.597 \pm 0.004$ & $0.091 \pm 0.001$\\
            & $[\mathsf V,\mathsf D]_{\mathrm{slice},50d}$
            & $0.914 \pm 0.002$
            & $0.252 \pm 0.007$ & $0.598 \pm 0.009$ & $0.092 \pm 0.000$\\
            & $[\mathsf V,\mathsf E,\mathsf D]_{\mathrm{slice},50d}$
            & $0.922 \pm 0.011$
            & $0.242 \pm 0.012$ & $0.596 \pm 0.004$ & $0.090 \pm 0.002$\\
            & $[\mathsf E,\mathsf D]_{\mathrm{Sobol},50d}$
            & $0.904 \pm 0.010$
            & $0.264 \pm 0.022$ & $0.611 \pm 0.007$ & $0.097 \pm 0.003$\\
        \cmidrule(lr){1-6}
        Separate probe
            & $\mathsf V_{\mathrm{slice},25d}\oplus\mathsf E_{\mathrm{Sobol},25d}$
            & $0.920 \pm 0.005$
            & $0.222 \pm 0.013$ & $0.631 \pm 0.003$ & $0.093 \pm 0.002$\\
            & $\mathsf V_{\mathrm{slice},25d}\oplus\mathsf D_{\mathrm{Sobol},25d}$
            & $0.914 \pm 0.003$
            & $0.240 \pm 0.015$ & $0.637 \pm 0.004$ & $0.094 \pm 0.002$\\
            & $\mathsf V_{\mathrm{slice},25d}\oplus[\mathsf E,\mathsf D]_{\mathrm{Sobol},25d}$
            & $0.918 \pm 0.003$
            & $0.230 \pm 0.013$ & $0.655 \pm 0.005$ & $0.095 \pm 0.002$\\
    
        \midrule
        \multicolumn{6}{l}{\textbf{AOCC--LPO}}\\
        Standalone
            & $\mathsf V_{\mathrm{slice},50d}$
            & $0.856 \pm 0.009$
            & $0.198 \pm 0.006$ & $0.301 \pm 0.005$ & $0.095 \pm 0.002$\\
            & $\mathsf E_{\mathrm{Sobol},50d}$
            & $0.876 \pm 0.015$
            & $0.213 \pm 0.013$ & $0.293 \pm 0.019$ & $0.109 \pm 0.002$\\
            & $\mathsf D_{\mathrm{Sobol},50d}$
            & $0.849 \pm 0.009$
            & $0.302 \pm 0.018$ & $0.273 \pm 0.008$ & $0.093 \pm 0.004$\\
        \cmidrule(lr){1-6}
        Shared probe
            & $[\mathsf V,\mathsf E]_{\mathrm{slice},50d}$
            & $0.902 \pm 0.003$
            & $0.230 \pm 0.006$ & $0.334 \pm 0.007$ & $0.097 \pm 0.003$\\
            & $[\mathsf V,\mathsf D]_{\mathrm{slice},50d}$
            & $0.890 \pm 0.003$
            & $0.224 \pm 0.010$ & $0.338 \pm 0.003$ & $0.102 \pm 0.003$\\
            & $[\mathsf V,\mathsf E,\mathsf D]_{\mathrm{slice},50d}$
            & $0.901 \pm 0.006$
            & $0.222 \pm 0.007$ & $0.339 \pm 0.009$ & $0.099 \pm 0.006$\\
            & $[\mathsf E,\mathsf D]_{\mathrm{Sobol},50d}$
            & $0.890 \pm 0.010$
            & $0.231 \pm 0.008$ & $0.277 \pm 0.010$ & $0.093 \pm 0.004$\\
        \cmidrule(lr){1-6}
        Separate probe
            & $\mathsf V_{\mathrm{slice},25d}\oplus\mathsf E_{\mathrm{Sobol},25d}$
            & $0.923 \pm 0.008$
            & $0.209 \pm 0.021$ & $0.319 \pm 0.005$ & $0.098 \pm 0.004$\\
            & $\mathsf V_{\mathrm{slice},25d}\oplus\mathsf D_{\mathrm{Sobol},25d}$
            & $0.869 \pm 0.008$
            & $0.281 \pm 0.022$ & $0.310 \pm 0.015$ & $0.102 \pm 0.001$\\
            & $\mathsf V_{\mathrm{slice},25d}\oplus[\mathsf E,\mathsf D]_{\mathrm{Sobol},25d}$
            & $0.899 \pm 0.023$
            & $0.240 \pm 0.039$ & $0.296 \pm 0.004$ & $0.100 \pm 0.011$\\
    
        \midrule
        \multicolumn{6}{l}{\textbf{relERT--LIO}}\\
        Standalone
            & $\mathsf V_{\mathrm{slice},50d}$
            & $0.863 \pm 0.004$
            & $0.231 \pm 0.023$ & $0.611 \pm 0.012$ & $1526.570 \pm 226.087$\\
            & $\mathsf E_{\mathrm{Sobol},50d}$
            & $0.866 \pm 0.009$
            & $0.323 \pm 0.019$ & $0.633 \pm 0.004$ & $2611.498 \pm 107.366$\\
            & $\mathsf D_{\mathrm{Sobol},50d}$
            & $0.880 \pm 0.006$
            & $0.313 \pm 0.011$ & $0.623 \pm 0.005$ & $2624.905 \pm 87.088$\\
        \cmidrule(lr){1-6}
        Shared probe
            & $[\mathsf V,\mathsf E]_{\mathrm{slice},50d}$
            & $0.881 \pm 0.006$
            & $0.283 \pm 0.003$ & $0.632 \pm 0.010$ & $2116.447 \pm 174.106$\\
            & $[\mathsf V,\mathsf D]_{\mathrm{slice},50d}$
            & $0.887 \pm 0.004$
            & $0.286 \pm 0.007$ & $0.636 \pm 0.005$ & $2304.207 \pm 96.243$\\
            & $[\mathsf V,\mathsf E,\mathsf D]_{\mathrm{slice},50d}$
            & $0.889 \pm 0.009$
            & $0.275 \pm 0.006$ & $0.635 \pm 0.004$ & $2219.269 \pm 27.328$\\
            & $[\mathsf E,\mathsf D]_{\mathrm{Sobol},50d}$
            & $0.865 \pm 0.012$
            & $0.320 \pm 0.023$ & $0.652 \pm 0.011$ & $2619.042 \pm 141.499$\\
        \cmidrule(lr){1-6}
        Separate probe
            & $\mathsf V_{\mathrm{slice},25d}\oplus\mathsf E_{\mathrm{Sobol},25d}$
            & $0.877 \pm 0.005$
            & $0.291 \pm 0.002$ & $0.675 \pm 0.001$ & $2527.422 \pm 95.422$\\
            & $\mathsf V_{\mathrm{slice},25d}\oplus\mathsf D_{\mathrm{Sobol},25d}$
            & $0.886 \pm 0.005$
            & $0.278 \pm 0.005$ & $0.684 \pm 0.003$ & $2605.108 \pm 12.768$\\
            & $\mathsf V_{\mathrm{slice},25d}\oplus[\mathsf E,\mathsf D]_{\mathrm{Sobol},25d}$
            & $0.887 \pm 0.005$
            & $0.270 \pm 0.005$ & $0.688 \pm 0.001$ & $2559.023 \pm 57.884$\\
    
        \midrule
        \multicolumn{6}{l}{\textbf{relERT--LPO}}\\
        Standalone
            & $\mathsf V_{\mathrm{slice},50d}$
            & $0.851 \pm 0.008$
            & $0.283 \pm 0.010$ & $0.435 \pm 0.003$ & $-2819.826 \pm 528.660$\\
            & $\mathsf E_{\mathrm{Sobol},50d}$
            & $0.812 \pm 0.009$
            & $0.455 \pm 0.023$ & $0.264 \pm 0.023$ & $-5344.462 \pm 1010.701$\\
            & $\mathsf D_{\mathrm{Sobol},50d}$
            & $0.800 \pm 0.006$
            & $0.445 \pm 0.011$ & $0.227 \pm 0.005$ & $-5749.645 \pm 121.957$\\
        \cmidrule(lr){1-6}
        Shared probe
            & $[\mathsf V,\mathsf E]_{\mathrm{slice},50d}$
            & $0.837 \pm 0.011$
            & $0.344 \pm 0.004$ & $0.425 \pm 0.010$ & $-2938.707 \pm 240.248$\\
            & $[\mathsf V,\mathsf D]_{\mathrm{slice},50d}$
            & $0.835 \pm 0.002$
            & $0.330 \pm 0.003$ & $0.427 \pm 0.008$ & $-2999.205 \pm 102.251$\\
            & $[\mathsf V,\mathsf E,\mathsf D]_{\mathrm{slice},50d}$
            & $0.846 \pm 0.006$
            & $0.314 \pm 0.004$ & $0.415 \pm 0.006$ & $-3370.628 \pm 235.862$\\
            & $[\mathsf E,\mathsf D]_{\mathrm{Sobol},50d}$
            & $0.832 \pm 0.008$
            & $0.410 \pm 0.005$ & $0.232 \pm 0.008$ & $-6025.267 \pm 1017.368$\\
        \cmidrule(lr){1-6}
        Separate probe
            & $\mathsf V_{\mathrm{slice},25d}\oplus\mathsf E_{\mathrm{Sobol},25d}$
            & $0.854 \pm 0.015$
            & $0.306 \pm 0.011$ & $0.455 \pm 0.005$ & $-2323.737 \pm 473.510$\\
            & $\mathsf V_{\mathrm{slice},25d}\oplus\mathsf D_{\mathrm{Sobol},25d}$
            & $0.871 \pm 0.013$
            & $0.331 \pm 0.013$ & $0.394 \pm 0.002$ & $-2457.306 \pm 347.344$\\
            & $\mathsf V_{\mathrm{slice},25d}\oplus[\mathsf E,\mathsf D]_{\mathrm{Sobol},25d}$
            & $0.874 \pm 0.009$
            & $0.295 \pm 0.011$ & $0.403 \pm 0.013$ & $-2742.574 \pm 691.268$\\
    
        \bottomrule
    \end{tabular}
    \label{tab:rq1_selection_behavior}
\end{table*}

\subsection{Selection behavior underlying RQ1}
\label{app:rq1_selection_behavior}

Table~\ref{tab:rq1_selection_behavior} describes the selection policies underlying the RQ1 outcome distributions. Selection entropy is $-\sum_{a\in\mathcal A}p_a\log p_a$, where $p_a$ is the empirical selection frequency of algorithm $a$; SBS and VBS agreement are the proportions of cases on which the selector chooses the fold-training SBS and the case-wise VBS, respectively. Conditional switch gain measures the mean native-scale improvement over the SBS among cases on which the selector chooses another algorithm:
$$
\begin{aligned}
    G_{\mathrm{switch}}
    &= \mathbb E\!\left[g_i\mid \hat a_i\neq a^{\mathrm{SBS}}_{j(i)}\right], \\
    g_i &=
    \begin{cases}
        y_{i,\hat a_i}-y_{i,a^{\mathrm{SBS}}_{j(i)}}, & \text{AOCC},\\
        y_{i,a^{\mathrm{SBS}}_{j(i)}}-y_{i,\hat a_i}, & \text{relERT},
    \end{cases}
\end{aligned}
$$
so positive values always indicate improvement. Entries are means $\pm$ standard deviations over three training seeds.

The diagnostics support the two mechanisms already visible in Fig.~\ref{fig:main_results}. Under LIO, separate-probe fusion raises VBS agreement while retaining positive switch gains, so its advantage arises from more useful case-wise departures within the same evaluation budget. Under relERT--LPO, every $\mathsf V$-containing selector has lower SBS agreement, higher VBS agreement, and a smaller conditional switch loss than the $\mathsf E$/$\mathsf D$-only selectors. Its improved central distribution therefore reflects genuine instance discrimination. Conditional switch gain nevertheless remains negative for every representation: a smaller set of costly mis-selections preserves the long upper tail and drives mean GC below zero.

\subsection{RQ3 analysis details}
\label{app:rq3}

\paragraph*{Linear-probe protocol}

For each outer fold and training seed, the ridge probe is fitted using training and test representations from the same frozen GeoPAS encoder, with the held-out predictions pooled across folds for reporting. Since $\mathsf V$ is learned under solver-performance supervision whereas $\mathsf E$ and $\mathsf D$ are fixed descriptors, the experiment aims to measure information accessibility after the prescribed training.

Let $F$ be the set of all folds and $\bar y_{a,\mathcal X_{\text{train}_i}}$ be the training mean for solver $a$. Its held-out prediction skill is
$$
\begin{aligned}
    S_a
    &:=
    1-
    \frac{\sum_{j=1}^{|F|}\sum_{i\in\mathcal X_{\text{test}_j}}(y_{ia}-\widehat y_{ja})^2}
         {\sum_{j=1}^{|F|}\sum_{i\in\mathcal X_{\text{test}_j}}(y_{ia}-\bar y_{a,\mathcal X_{\text{train}_j}})^2},\\
    S&:=\frac{1}{|\mathcal A|}\sum_{a\in\mathcal A}S_a.
\end{aligned}
    \label{eq:rq3-prediction-skill}
$$
Thus $S_a=0$ corresponds to the fold-training-mean predictor; $S$ is the macro mean over the seven solvers. Here $y$ is native AOCC or $\log(\mathrm{relERT})$. Solver choices are evaluated on native AOCC or relERT using the training-fold SBS, with all case scores pooled before GC is computed.

Ridge inputs are standardized using outer-training statistics, with non-finite values median-imputed and constant coordinates removed. To prevent near-inactive learned coordinates from being amplified, the standard deviations of \(\mathsf V\) and \(\mathsf D\) are floored at \(10^{-3}\) times the median positive within-block standard deviation. Concatenated blocks are divided by the square root of their retained widths. Ridge regression uses \(\lambda=10^{-2}\) and an unpenalized intercept; scale-floor factors \(10^{-4}\) and \(10^{-2}\) yield the same conclusions.

For the auxiliary solver-contrast task, we use
\begin{equation}
    \widetilde y_{ia}
    :=
    y_{ia}-\frac{1}{|\mathcal A|}\sum_{b\in\mathcal A}y_{ib}.
    \label{eq:rq3-contrast}
\end{equation}
Because ridge regression is linear in the target matrix and uses the same penalty for every solver, the corresponding fitted predictions equal the row-centered absolute-target predictions. Their solver ordering is therefore unchanged, so selection is reported once (Figure~\ref{fig:rq3-contrast}).

\begin{figure}[t]
    \centering
    \includegraphics[width=\linewidth]{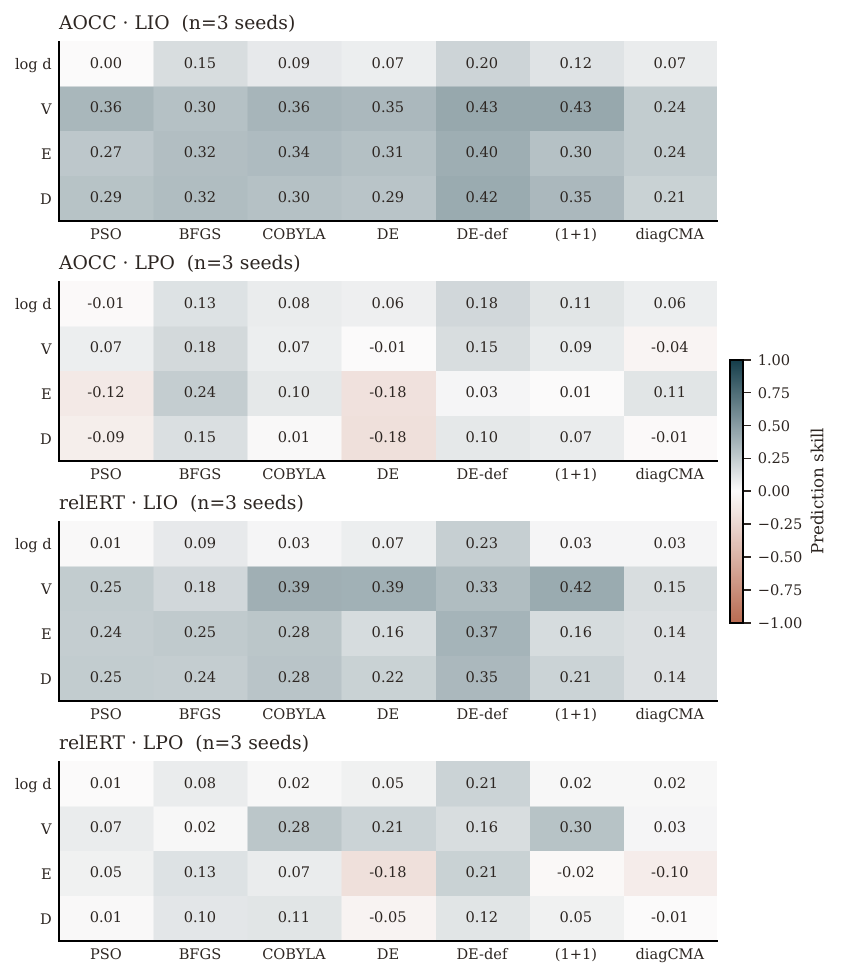}
    \caption{Per-solver prediction skill for the centered targets in Eq.~\eqref{eq:rq3-contrast}, averaged over three training seeds. The four rows in each panel are the dimension-only, $\mathsf V$, $\mathsf E$, and $\mathsf D$ probes. }
    \label{fig:rq3-contrast}
\end{figure}


\paragraph*{Representation overlap}

Linear centered-kernel alignment (CKA) is computed separately within each held-out fold after column standardization~\cite{kornblith2019similarity}. We report the unadjusted value, the value after subtracting dimension means, and, for LIO, the value after subtracting means within each dimension--BBOB-function-group cell. Fold values are first averaged within each seed and then across the three seeds. Under LPO each fold contains one function, so function-group adjustment is identical to dimension adjustment and is not reported separately.

\begin{figure}[t]
    \centering
    \includegraphics[width=\linewidth]{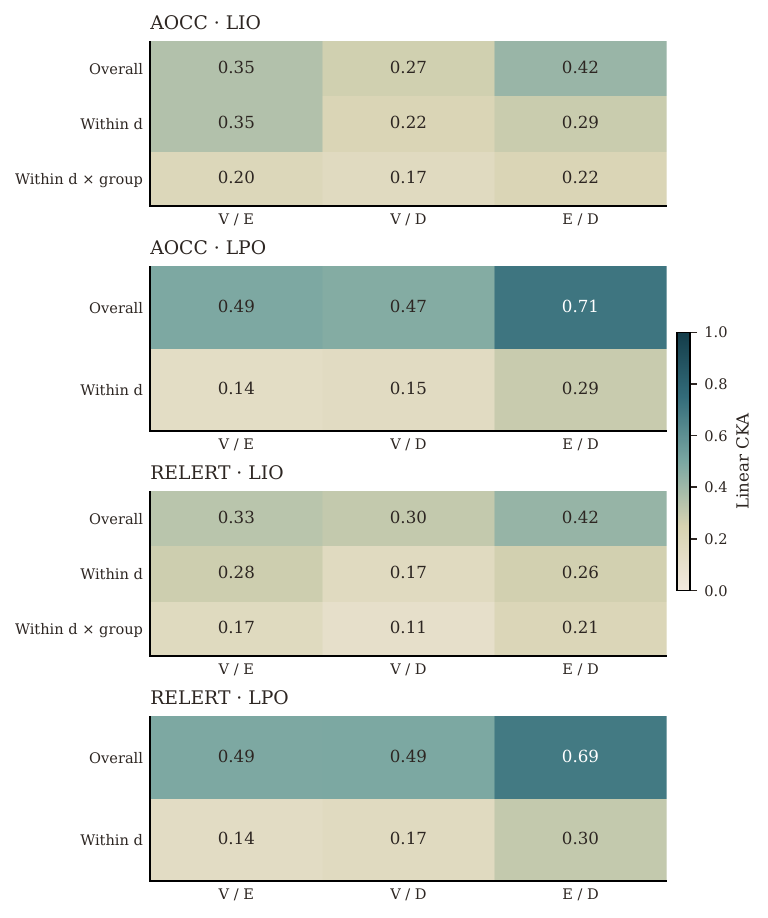}
    \caption{Fold-local linear CKA between the frozen representation pairs. ``Within $d$'' removes dimension means, and ``within $d\times$ group'' additionally removes broad BBOB function-group means and is reported only for LIO. Cells average folds within seeds and then the three training seeds.}
    \label{fig:rq3-cka}
\end{figure}

As in Figure~\ref{fig:rq3-cka}, the overall CKA of $\mathsf V$ with $\mathsf E$ or $\mathsf D$ ranges from $0.266$ to $0.349$ under LIO and from $0.472$ to $0.494$ under LPO. After removal of dimension and function-group means, the corresponding LIO values are $0.110$--$0.201$; after dimension-mean removal, the LPO values are $0.141$--$0.168$. The stronger unadjusted LPO alignment therefore reflects, in substantial part, common dimension-level organization within the held-out problems. The remaining alignment is nonzero, but lower CKA is not itself evidence that either representation adds solver-relevant predictive information, as shown in Figure~\ref{fig:rq3-incremental}.

}
 
%


 




\vfill

\end{document}